\documentclass[runningheads]{llncs}
\usepackage{graphicx}

\usepackage{tikz}
\usepackage{comment}
\usepackage{amsmath,amssymb} 

\usepackage[accsupp]{axessibility}  

\usepackage{url}            
\usepackage{booktabs}       
\usepackage{nicefrac}       
\usepackage{microtype}      
\usepackage{xcolor}         

\usepackage[T1]{fontenc}
\usepackage[latin1]{inputenc}
\usepackage{cite}

\usepackage{amsmath}

\usepackage{lipsum, booktabs}

\usepackage{algorithmicx}
\usepackage{algorithm}
\usepackage[noend]{algpseudocode}
\usepackage{mathtools}

\newcommand{\figref}[1]{Fig.~\ref{#1}}
\newcommand{\tabref}[1]{Tab.~\ref{#1}}
\newcommand{\secref}[1]{Sec.~\ref{#1}}

\newcommand{\dnet}{DomainNet}

\usepackage{float}
\usepackage{soul}
\usepackage{multirow}
\usepackage{wrapfig}
\usepackage{IEEEtrantools}

\usepackage{booktabs}

\usepackage{subcaption}
\usepackage[toc,page]{appendix}

\newcommand{\Ours}{MemSAC}

\newcommand{\E}{\mathcal{E}}
\newcommand{\F}{\mathcal{F}}
\newcommand{\C}{\mathcal{C}}
\newcommand{\LL}{\mathcal{L}}

\newcommand{\B}{\mathcal{B}}
\newcommand{\G}{\mathcal{G}}
\newcommand{\Y}{\mathcal{Y}}
\newcommand{\X}{\mathcal{X}}
\newcommand{\M}{\mathcal{M}}

\newcommand{\shortpara}[1]{\noindent {\bf {#1}}  \hspace{4pt}}

\newcommand{\N}{\mathcal{N}}

\usepackage{pifont}
\newcommand{\tick}{{{{\ding{51}}}}}
\newcommand{\cross}{{{{\ding{55}}}}}
\usepackage[inline]{enumitem}

\newenvironment{tight_enumerate}{
\begin{enumerate}[leftmargin=15pt]
  \setlength{\topsep}{0pt}
  \setlength{\itemsep}{0pt}
  \setlength{\parskip}{0pt}
  \setlength{\parsep}{0pt}
}{\end{enumerate}}

\usepackage[pagebackref=true,breaklinks=true,colorlinks=true,bookmarks=false,citecolor=blue]{hyperref}
\usepackage[capitalise,noabbrev,nameinlink]{cleveref}
\usepackage[english]{babel}

\begin{document}
\pagestyle{headings}
\mainmatter
\def\ECCVSubNumber{3093}  

\title{\Ours{}: Memory Augmented Sample Consistency for Large Scale Domain Adaptation } 

\titlerunning{MemSAC: Large Scale Domain Adaptation}
%
\author{Tarun Kalluri \and
Astuti Sharma \and
Manmohan Chandraker \\
}
\authorrunning{T. Kalluri et al.}
%
\institute{University of California San Diego, La Jolla CA 92093, USA \\ 
\email{\{sskallur,asharma,mkchandraker\}@eng.ucsd.edu}
}
\maketitle

\begin{abstract}

Practical real world datasets with plentiful categories introduce new challenges for unsupervised domain adaptation like small inter-class discriminability, that existing approaches relying on domain invariance alone cannot handle sufficiently well. In this work we propose \Ours{}, which exploits sample level similarity across source and target domains to achieve discriminative transfer, along with architectures that scale to a large number of categories. For this purpose, we first introduce a memory augmented approach to efficiently extract pairwise similarity relations between labeled source and unlabeled target domain instances, suited to handle an arbitrary number of classes. Next, we propose and theoretically justify a novel variant of the contrastive loss to promote local consistency among within-class cross domain samples while enforcing separation between classes, thus preserving discriminative transfer from source to target. We validate the advantages of \Ours{} with significant improvements over previous state-of-the-art on multiple challenging transfer tasks designed for large-scale adaptation, such as \dnet{} with 345 classes and fine-grained adaptation on Caltech-UCSD birds dataset with 200 classes. We also provide in-depth analysis and insights into the effectiveness of \Ours{}. Code is available on the project webpage  \href{https://tarun005.github.io/MemSAC}{https://tarun005.github.io/MemSAC}.
\end{abstract}

\section{Introduction}
\label{sec:intro}

It is well known that deep neural networks often do not generalize well when the distribution of test samples differ significantly from those in training. Unsupervised domain adaptation seeks to improve transferability in the presence of such domain shift, for which a variety of approaches have been proposed \cite{ben2010theory, ben2006analysis, DANN, CDAN, long2013transfer, long2015learning, long2016unsupervised, bousmalis2016domain, bousmalis2017unsupervised, saito2017adversarial, tzeng2014deep, tzeng2015simultaneous, tzeng2017adversarial, xu2019larger, chen2019transferability, gu2020spherical, na2021fixbi, ducross21}. Despite impressive gains, most approaches have been largely demonstrated on datasets with a limited number of categories \cite{saenko2010adapting, peng2017visda}, 

\begin{figure}[t]
\begin{center}
    \begin{minipage}[b]{0.35\textwidth}
        \centering
        \includegraphics[width=\textwidth]{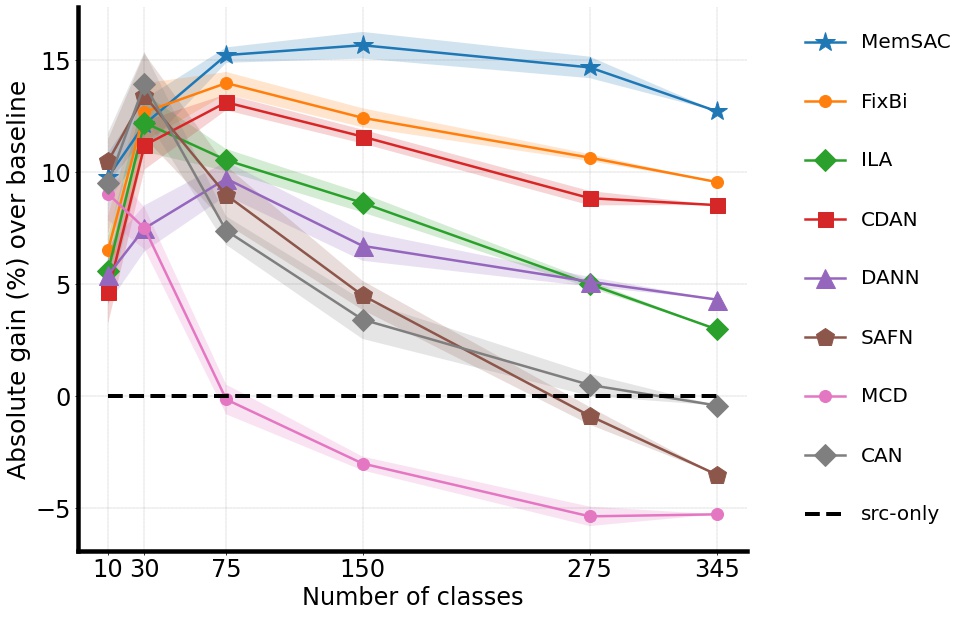}
        \vspace{-8pt}
        \subcaption{}
        \label{fig:class_vs_acc}
    \end{minipage}
    \hfill
    \begin{minipage}[b]{0.6\textwidth}
        \centering
        \includegraphics[width=\textwidth]{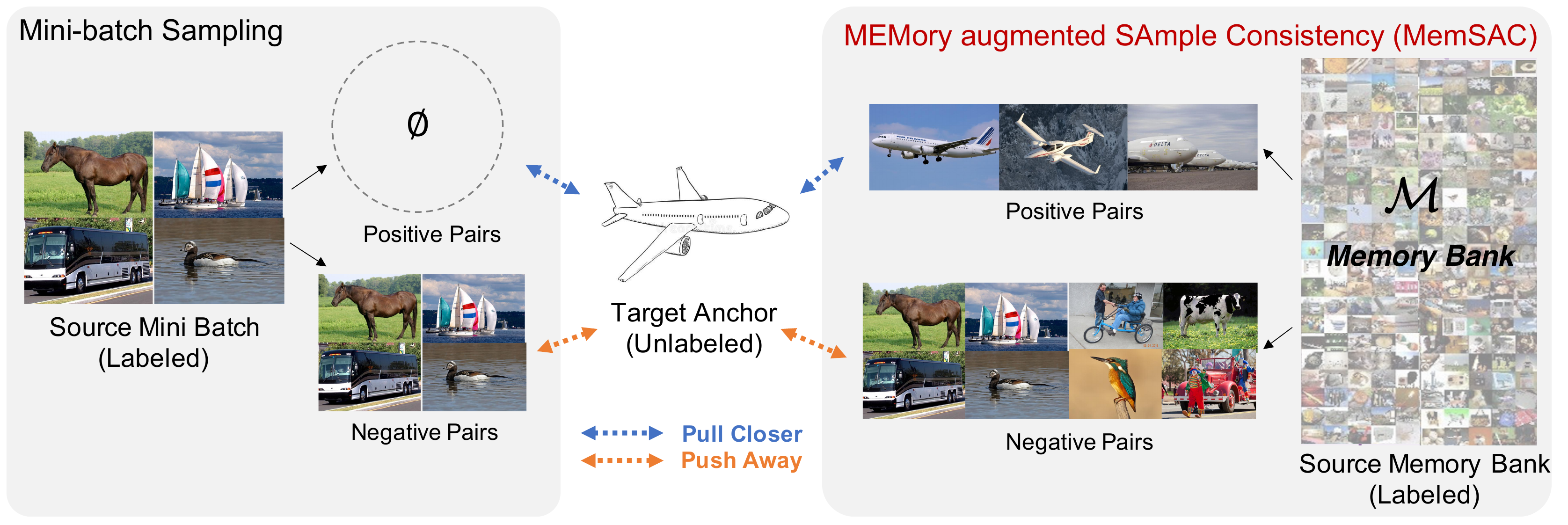}
        \subcaption{}
    \label{fig:intro_pic}
    \end{minipage}
    \captionsetup{width=\textwidth, font=footnotesize}
    \caption{(\subref{fig:class_vs_acc}) Accuracy(\%) of various methods proposed for unsupervised domain adaptation with respect to the number of training classes from \dnet{}\cite{peng2019moment}(\textbf{R}{$\rightarrow$}\textbf{C}). While most methods perform equally well for smaller number of categories (10-30), the benefits diminish with increasing number of classes in the dataset, to the extent that the performance drops \emph{even below the source-only baseline} for few methods. In contrast, proposed \Ours{} obtains significant gains ($\sim15\%$) even on large scale datasets with many classes~\cite{peng2019moment}. (\subref{fig:intro_pic}) {\bf MEMory augmented SAmple Consistency (\Ours{})} The proposed method uses a memory bank and a sample consistency loss to identify source samples across a large number of categories that likely belong to the same class as an unlabeled target example, then pulls them together in feature space while pushing away samples from all other classes. Notice that without the proposed feature aggregation, a target anchor sample might not find any positive pairs ($\emptyset$) leading to noisy consistency estimates.}
    \vspace{-16pt}
\end{center}
\end{figure}

We first ask the question of whether existing domain adaptation methods scale to a large number of categories. Surprisingly, the answer is usually no. To illustrate this, consider Figure \ref{fig:class_vs_acc}, which plots the absolute gain over a source-only model obtained by well-known adaptation methods (including DANN~\cite{DANN}, MCD~\cite{saito2018maximum}, SAFN~\cite{xu2019larger}, CAN~\cite{kang2019contrastive}, FixBi~\cite{na2021fixbi}) with respect to number of classes sampled from the \dnet{} dataset \cite{peng2019moment}. While all methods provide similar benefits over a source-only model in smaller-scale settings with 10-30 classes, the gains reduce significantly when faced with a few hundred classes, where accuracies may even become {\em worse than a source-only model}. 

We postulate that the above limitations with a larger number of categories arise due to lower inter-class separation and a greater possibility of negative transfer. Our key design choices stem from simple yet effective mechanisms developed in other areas such as self-supervised learning that can significantly benefit many-class domain adaptation. The resulting method, \Ours{} (MEMory augmented SAmple Consistency), achieves impressive performance gains to establish new state-of-the-art on datasets such as DomainNet (345 classes) and CUB (200 classes). In the same illustration above, \Ours{} obtains large improvements of 14.6\% over a source-only baseline for 275 classes and 12.7\% for 345 classes. 

Our first insight for many-class domain adaptation pertains to class confusion, where several classes possibly look similar to each other. Classical adversarial approaches~\cite{DANN, saito2018maximum, xie2018learning, xu2019larger, kang2019contrastive} which rely on domain alignment alone do not acknowledge this, giving rise to negative transfer as two seemingly close classes might align with each other. This problem is exacerbated in the extreme case of fine-grained datasets, where all the classes look similar to each other. On the other hand, class specific alignment strategies \cite{na2021fixbi, ducross21, saito2017adversarial, pei2018multi, kang2019contrastive, na2021fixbi} suffer from noisy psuedo-labels leading to poor transfer. We observe that the contrastive loss is shown to be highly successful in learning better transferable features \cite{he2020momentum, chen2020improved, chen2020simple, wu2018unsupervised, henaff2019data, misra2020self, gordon2020watching, grill2020bootstrap} and seek to extend those benefits to many-class domain adaptation. We achieve this with a novel {\em cross-domain sample consistency} loss which tries to align each sample in source domain with related samples in target domain, achieving tighter clusters and improved adaptation in the process. We provide theoretical justification for the effectiveness of our proposed loss by showing that it is akin to minimizing an upper bound to the input-consistency regularization recently proposed in \cite{wei2020theoretical}, thereby ensuring that locally consistent prediction provides accuracy guarantees on unlabeled target data for unsupervised domain adaptation.

Our second insight pertains to architectural choices for training with a large number of categories. While having access to plentiful positive and negative pairwise relations per training iteration is desirable to infer local structure, the number of possible pairs are inherently restricted by the batch-size which is in turn limited by the GPU memory. We efficiently tackle this challenge in \Ours{} by augmenting the adaptation framework with a lightweight, non-parametric memory module. Distinct from prior works~\cite{he2020momentum, wang2020cross}, the memory module in our setting aggregates the \textit{labeled} source domain features from multiple recent mini-batches, thus providing \textit{unlabeled} target domain anchors meaningful interactions from sizeable positive and negative pairs even with reasonably small batch sizes that do not incur explosive growth in memory (\figref{fig:intro_pic}). Our architecture scales remarkably well with the number of categories, including the case of fine-grained adaptation~\cite{PAN} where all classes belong to a single {subordinate category} \cite{zhang2014part, branson2014bird}. Moreover, \Ours{} incurs negligible overhead in terms of speed and GPU memory during training and testing, making it an attractive choice for real-world usage of large-scale adaptation.

To summarize, in contrast to prior works, \Ours{} achieves scalability in domain adaptation with a large number of classes. Our main contributions are:
\begin{tight_enumerate}
\item A novel cross-domain sample consistency loss to enforce closer clustering of same category samples across source and target domains by exploiting pairwise relationships, thus achieving improved domain transfer even with many categories (\secref{subsec:crdoco}). 
\item A memory-based mechanism to handle limited batch-sizes by storing past features and effectively extracting similarity relations over a larger context for large scale datasets (\secref{subsec:memory}). 
\item Theoretical justification of the proposed losses in terms of the input-consistency regularization proposed in \cite{wei2020theoretical} for domain adaptation (\secref{subsec:discussion}).
\item A new state-of-the-art that outperforms all prior approaches by a significant margin on datasets with a large number of categories, such as $4.02\%$ and $4.65\%$ improvements in accuracy over the baseline which does not use our loss on the challenging \dnet{} dataset with 345 categories and CUB-Drawings with 200 categories, respectively (\secref{sec:expmnts}).
\end{tight_enumerate}

\section{Related Work}

\noindent {\bf Unsupervised Domain Adaptation}
A suite of tools have been proposed recently under the umbrella of unsupervised domain adaptation (UDA) that enable training on a labeled source domain and deploy models on a different target domain with few or no labels. 
A large body of these works aim to minimize some notion of divergence~\cite{ben2006analysis, ben2010theory, saenko2010adapting} between the source and target using an adversarial objective, resulting in domain invariant features~\cite{DANN, tzeng2014deep, tzeng2017adversarial, xie2018learning, CDAN, tzeng2015simultaneous, chen2019progressive, saito2018maximum, kalluri2019universal}. 
Since domain invariance alone does not guarantee discriminative features in target~\cite{kumar2018co}, recent approaches propose class aware adaptation to align class conditional distributions across source and target~\cite{saito2017asymmetric, pei2018multi, xie2018learning, kang2019contrastive, ILA, na2021fixbi, ducross21, wei2021toalign, cui2020hda}. ATT~\cite{saito2017asymmetric} assigns pseudo-labels based on predictions from classifiers, MADA~\cite{pei2018multi} uses separate adversarial networks for each class, ILA~\cite{ILA} computes pairwise similarity between samples within a mini-batch for instance aware adaptation while SAFN~\cite{xu2019larger} proposes re-normalizing features to achieve transferability. However, none of these works explicitly address the issue of scalability to adaptation with a large number of categories. Moreover, many clustering based methods~\cite{kang2019contrastive, park2020joint} and instance based methods~\cite{ILA, wang2021cross} proposed for UDA are not readily scalable to large datasets.

While partial adaptation~\cite{cao2018partial, cao2018partialSAN, zhang2018importance}, open set adaptation~\cite{panareda2017open, saito2018open} and universal adaptation~\cite{saito2020universal} allow training on real world source datasets with many categories, they are only focused on adaptation across those categories that are shared between source and target which are generally few in number, and do not address the problem of discriminative transfer across \textit{all} the categories which is a different practical problem, and focus of this work.

\noindent {\bf Fine Grained Domain Adaptation}
Fine grained visual categorization deals with classifying images that belong to a single subordinate category, such as birds, trees or animal species~\cite{WahCUB_200_2011, inat2017}. While fine grained classification on within domain samples has received much attention~\cite{zhang2012pose, zheng2017learning, sun2018multi, zhang2014part, branson2014bird, lin2015bilinear, zheng2019looking}, the problem of unsupervised domain adaptation across fine-grained categories is relatively less studied \cite{gebru2017fine, cui2018large, xu2016webly, PAN}. All prior works often demand additional annotations in the form of attributes~\cite{gebru2017fine}, weak supervision~\cite{cui2018large}, part annotations~\cite{xu2016webly} or hierarchical relationships~\cite{PAN} in one of the domains which might not be universally available. In contrast, we propose a method that performs fine-grained adaptation requiring no such additional knowledge. 

\noindent {\bf Contrastive Learning} 
The success of contrastive learning~\cite{hadsell2006dimensionality, gutmann2010noise, arora2019theoretical, wei2020theoretical} in extracting visual representations from data has attracted wide interest in self-supervised \cite{he2020momentum, chen2020improved, chen2020simple, wu2018unsupervised, henaff2019data, misra2020self, gordon2020watching, grill2020bootstrap, cole2021does}, semi-supervised~\cite{assran2021semi} and supervised learning~\cite{khosla2020supervised}. A unifying idea in those works is to encourage positive pairs, which are often augmented versions of the same image, to have similar representations in the feature space while pushing negative pairs far away. {However, all those prior works assume that all positive and negative pairs in the contrastive loss come from the same domain. In contrast, we propose a variant of contrastive loss to handle multi-class discriminative transfer by enforcing sample consistency across similar samples extracted from different domains. }

\section{Unsupervised Adaptation using \Ours{}}

\begin{figure*}[t]
\begin{center}
        \centering
        \includegraphics[width=0.8\textwidth]{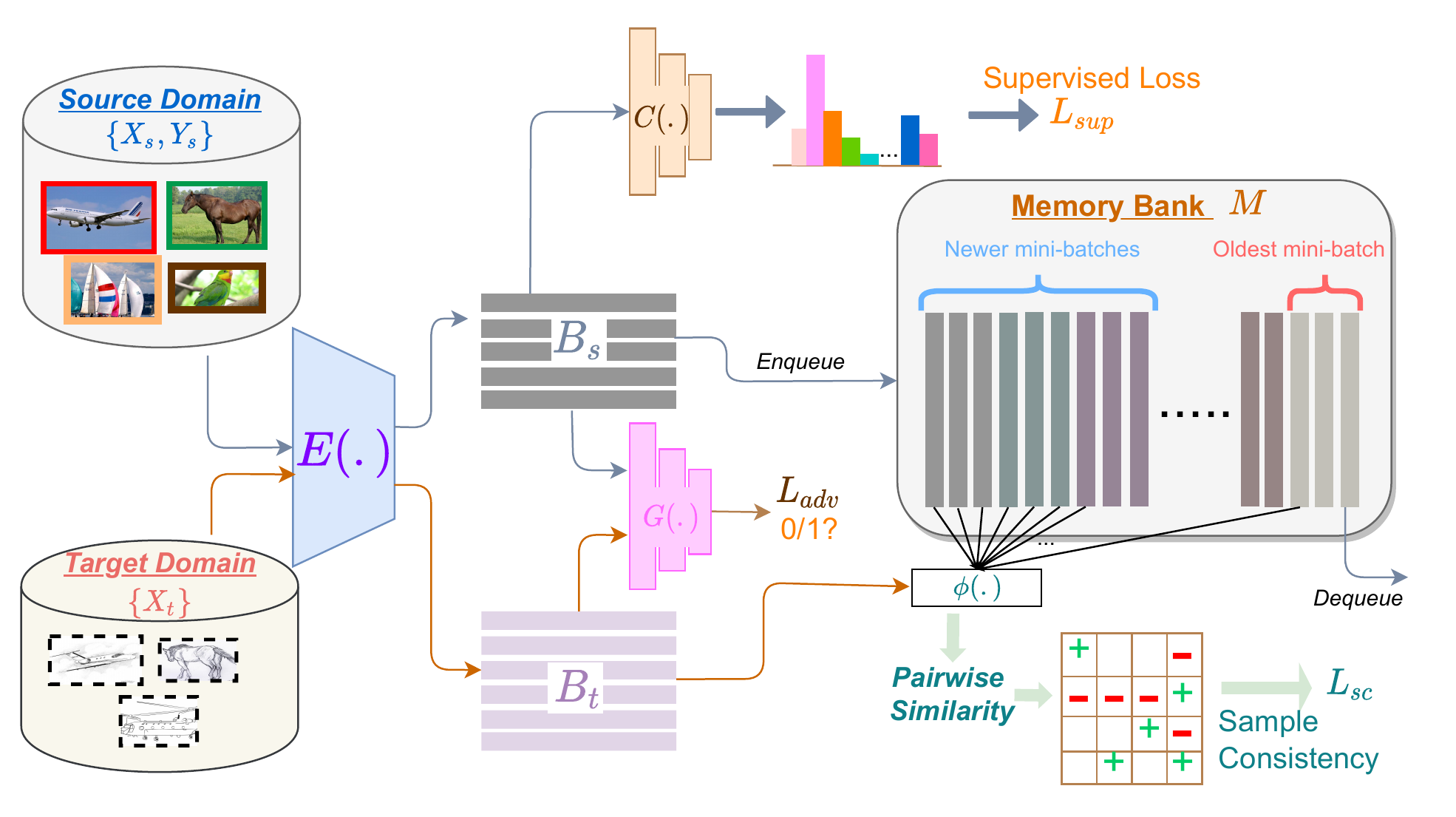}
        \captionsetup{width=\textwidth, font=footnotesize}
        \caption{{\bf An overview of \Ours{} for domain adaptation} During each iteration, the 256-dim source feature embeddings computed using $\E$, along with their labels, are added to a memory bank $\M$ and the oldest set of features are removed. Pairwise similarities between each target feature in mini-batch and all source features in memory bank are used to extract possible within-class and other-class source samples from the memory bank. Using the proposed consistency loss ($\LL_{sc}$) on these similar and dissimilar pairs, along with adversarial loss ($\LL_{adv}$), we achieve both local alignment and global adaptation.
        }
        \label{fig:model_pic}
\end{center}
\vspace{-16pt}
\end{figure*}

{\bf Problem Description} In unsupervised domain adaptation, we have {labeled} samples $\X^s$ from a source domain with a corresponding {source} probability distribution $P_s$, labeled according to a {true} labeling function $f^*$, and $\Y^s = f^*(\X^s)$. We are also given {unlabeled} data points $\X^t$ sampled according to the {target} distribution $P_t$. We follow a \textit{covariate shift assumption}~\cite{ben2010theory}, where we assume that the marginal source and target distributions $P_s$ and $P_t$ are different, while the {true} labeling function $f^*$ is same across the domains. The labels belong to a fixed category set $\Y=\{1,2,\ldots,C\}$ with $C$ different categories. 
Provided with this information, the goal of any learner is to output a predictor that achieves good accuracy on the target data $\X_t$. A key novelty in our instantiation of this framework lies in proposing an adaptation approach that works well even with a large number of classes $C$, by efficiently handling class confusion and discriminative transfer. The overview of the proposed architecture is shown in \figref{fig:model_pic}. $\E$ and $\C$ are the feature extractor and the classifier respectively. The objective function for \Ours{} is given by
\begin{IEEEeqnarray}{C}
    \min_{\theta} \LL_{sup}(\X^s, \Y^s ; \theta) + \lambda_{adv} \LL_{adv}(\X^s,\X^t ; \theta)
    + \lambda_{sc} \LL_{sc}(\X^s, \Y^s, \X^t ; \theta),
    \label{eq:total_loss}
\end{IEEEeqnarray}
\noindent where $\LL_{sup}$ is the supervised loss on source data, or the cross-entropy loss between the predicted class probability distributions and ground truths computed on source data. $\LL_{adv}$ is the domain adversarial loss which we implement using a class conditional discriminator (Eq.~\ref{eq:ladv}) and $\LL_{sc}$ is our novel cross-domain sample-consistency loss which is used to enforce the local similarity (or dissimilarity) between samples from source and target domains (Eq.~\ref{eq:sc_loss}). $\lambda_{adv}$ and $\lambda_{sc}$ are the corresponding loss coefficients. We use $\B_s( \in \X^s)$ and $\B_t( \in \X^t)$ to denote labeled source and unlabeled target mini-batches respectively, which are chosen randomly at each iteration from the dataset. 

\noindent {\bf Class conditional adversarial loss} We adopt the widely used adversarial strategy to learn domain-invariant feature representations using a domain discriminator $\G(.,\omega)$ parametrized by $\omega$.
To address the novel challenges presented by the current setting with large number of classes, we adopt the multilinear conditioning proposed in CDAN~\cite{CDAN} to fuse information from the deep features as well as the classifier predictions. Denoting $f = \E(x)$ and $g = \C(\E(x))$, the input $h(x)$ to the discriminator $\G$ is given by
$
    h(x) = T_{\otimes} (g , f )(x)  = f(x) \otimes g(x),
$
where $\otimes$ refers to the multilinear product (or flattened outer product) between the feature embedding and the softmax output of the classifier.
The discriminator and adversarial losses are then computed as
\begin{IEEEeqnarray}{C}
    \LL_{d} = \frac{1}{|\B_s|} \sum_{i \in \B_s}\!\!\!\! - \log (\G(h_i; \omega)) + \frac{1}{|\B_t|} \sum_{i \in \B_t}\!\!\!\! - \log (1-\G(h_i; \omega)) \qquad
    \LL_{adv} = - \LL_d. \IEEEeqnarraynumspace
    \label{eq:ladv}
\end{IEEEeqnarray}
We note that our contributions are complementary to the type of alignment objective used. In \tabref{tab:dnet+dann}, we show significant gains starting from another adversarial objective (DANN \cite{DANN}) and MMD objectives (CAN ~\cite{kang2018deep}) as well.

\subsection{Cross domain sample consistency} 
\label{subsec:crdoco}

To achieve category specific transfer from source to target, we propose using much finer sample-level information to enforce consistency between similar samples, while also separating dissimilar samples across domains.
Since our final goal is to transfer the class discriminative capability from source to target, we define the notions of similarity and dissimilarity as follows. For each target sample $x_t$ from a target mini-batch $\B_t$ as the anchor, we construct a \textit{similar set} $\B_{s^+}^{x_t} = \{ x \in\B_s | f^*(x)=f^*(x_t) \}$ and dissimilar set $\B_{s^-}^{x_t} {=} \B_s{\setminus}{\B_{s^+}^{x_t}}$ consisting of source samples and use this knowledge of sample-level similarity in the following \textit{sample consistency loss}
\begin{IEEEeqnarray}{rCl}
    \LL_{sc,\B} &=& \frac{1}{|\B_t|} \sum_{j \in \B_t} -\log \left\{ \sum_{i \in \B_{s^+}^j} \frac{\exp(\phi_{ij}/\tau)}
                {\sum_{i \in \B_s} \exp(\phi_{ij}/\tau) }  \right\} \IEEEeqnarraynumspace
    \label{eq:sc_loss_batch}
\end{IEEEeqnarray}          
where $\phi_{ij}$ measures the cosine similarity metric between two feature vectors $i$ and $j$, ($\phi_{ij} {=} \phi(f_i , f_j) {=} \tfrac{f_i \cdot f_j}{||f_i||||f_j||}$) and $\tau$ is the temperature parameter used to scale the contributions of positive and negative pairs~\cite{chen2020simple, hinton2015distilling}. $\LL_{sc,\B}$ denotes the sample consistency loss computed using the mini-batch. Distinct from standard constrative loss~\cite{misra2020self, chen2020simple} that typically derives positive pairs from augmented versions of the same image, our loss in Eq.~\eqref{eq:sc_loss_batch} is well-suited to handle multiple positive and negative pairs for each anchor, similar to SupCon loss~\cite{khosla2020supervised}. However, in contrast to SupCon, our modified consistency loss allows us to scale domain adaptation to many-class settings.

\noindent \textbf{kNN-based pseudo-labeling} There are two challenges in directly using the sample consistency loss in \eqref{eq:sc_loss_batch}. Firstly, unlike prior approaches~\cite{henaff2019data, chen2020simple, misra2020self} that use random transformations of same image to construct positives and negatives, the target data in unsupervised domain adaptation is completely unlabeled, so we do not have the similarity information readily ($f^*(x_t)$ is unknown).
To address this issue, we use a k-NN based psuedo-labeling trick for all the target samples in a mini-batch. {In every iteration of the training, for each target sample $x_t$ from the target training mini-batch $\B_t$, we find $k$ nearest neighbors from the source training mini-batch $\B_s$, }which are computed using the feature similarity scores $\phi_{i,x_t}$. $x_t$ is then assigned the label corresponding to the majority class occurring among its neighbors. We use a value of $k{=}5$. 
Such an approach for psuedo-labeling is independent of, thus less sensitive to, noisy classifier boundaries helping us extract reliable target psuedo-labels during training. Once $\B_t$ is psuedo-labeled, it is straightforward to compute $\B_{s^+}^{x_t}$ in \eqref{eq:sc_loss_batch}. The second challenge is lack of representation for all classes in a mini-batch, which we address next.

\subsection{Memory augmented similarity extraction}
\label{subsec:memory}

From Eq.~\eqref{eq:sc_loss_batch}, we can observe that if the source and target mini-batches $\B_s$ and $\B_t$ contain completely non-intersecting classes, then the pseudo labeling of targets and the subsequent sample consistency loss would be noisy and lead to negative impact. This problem is exacerbated in our setting with a large number of classes, as randomly sampled $\B_s$ and $\B_t$ usually contain images with mutually non-intersecting categories. While one solution is to increase the size of mini-batch, it comes with significant growth in memory which is not scalable. 

Therefore, we propose using a non-parametric memory bank $\M$ that aggregates the computation-free features, along with the corresponding labels, across multiple past mini-batches from the source dataset. We note that if the size of the memory bank $|\M|$ is sufficiently large, then source samples from all the classes would be adequately present in $\M$, providing us with authentic positive and negative samples for use in the sample consistency loss. Furthermore, since the memory overhead of storing the features in the memory bank itself is negligible (we only store the computation-free features), proposed adaptation approach can be scaled to handle arbitrarily large number of classes, as datasets with larger classes only requires us to correspondingly increase the size of $\M$, thus decoupling the similarity computation with mini-batch size or dataset size. 
Different from prior approaches that augment training with memory module \cite{wu2018unsupervised,he2020momentum,wang2020cross}, our approach aggregates features from multiple source batches, thus helping target samples to extract meaningful pairwise relationships from different classes.

\noindent {\bf Initializing and updating memory bank } To initialize the memory bank, we first bootstrap the feature extractor for few hundred iterations by training only using $\LL_{sup}$ and $\LL_{adv}$ losses, and then introduce our consistency loss $\LL_{sc}$ and start populating $\M$. After this, 
we follow a queue based approach for updating the memory bank similar to XBM~\cite{wang2020cross}. In each iteration, We remove (\textit{dequeue}) the oldest batch of features from the queue and insert (\textit{enqueue}) the fresh mini-batch of source features (computed as $\{\E(x) | x \in \B_s\}$) along with the corresponding source labels. Alternative strategies for updating $\M$, such as a momentum encoder~\cite{he2020momentum}, yield similar results (refer \secref{sec:momentum_update} in the {supplementary}).

\vspace{2pt}
\noindent {\bf Sample consistency using memory bank } We can now use $\M$ as a proxy for $\B_s$ (and similar set $\M_+^{x_t}$ as a proxy for $\B_{s^+}^{x_t}$) in assigning the target psuedo labels and in the sample consistency loss in \eqref{eq:sc_loss_batch}. $|\M|$ is often much higher than $|\B_s|$, so access to larger number of source samples from $\M$ provides k-NN pseudo labels that are more reliable, with richer variety of positive and negative pairwise relations (refer \secref{sec:knnpseudo} in the {supplementary}). The final sample consistency loss used in \Ours{} is
\vspace{-0.2cm}
\begin{IEEEeqnarray}{c}
    \LL_{sc} = \frac{1}{|\B_t|} \sum_{j \in \B_t} -\log \left\{ \sum_{i \in \M_{+}^j} \frac{\exp(\phi_{ij}/\tau)}
                {\sum_{i \in \M} \exp(\phi_{ij}/\tau) }  \right\}. \IEEEeqnarraynumspace
    \label{eq:sc_loss}
\end{IEEEeqnarray}          

\subsection{Theoretical Insight}
\label{subsec:discussion}

%


%
Recently, Wei et al. \cite{wei2020theoretical} provide theoretical validation for contrastive learning. Specifically, under an \textit{expansion} assumption which states that class conditional distribution of data is locally continuous, they bound the target error of a classifier $C$ parametrized by $\theta$ by encouraging consistent predictions on neighboring examples. The regularization objective $R(\theta)$ is given by
$
    R(\theta) \equiv \min_{\theta} \mathbb{E}_x[\max_{x' \in \N(x)} \mathbf{1}(C(x; \theta) \neq C(x'; \theta))],
$
where $\N(x)$ is the neighborhood of a sample $x$ (Eq 1.2 in \cite{wei2020theoretical}). 
We now show the connections that can be drawn between our loss and the theory proposed in \cite{wei2020theoretical}.
For this purpose, we work with the following approximations. Firstly, we approximate the neighborhood $\N(x)$ of a sample $x$ with the \textit{similar set} defined in \secref{subsec:crdoco}, that is $\N(x) = \B_{+}^{x}$. 
Next, we approximate the hard condition that the classifier outputs of two images be equal $\mathbf{1}(C(x; \theta) \neq C(x'; \theta))$, with the soft probability $\mathbf{Pr}(C(x; \theta) \neq C(x'; \theta))$. Starting with the above objective, we have
%
%
\begin{IEEEeqnarray*}{rCl}
    && \max_{x' \in \N(x)} \mathbf{1}(C(x; \theta) \neq C(x'; \theta))   \\
    &\leq & \sum_{x' \in \N(x)} \mathbf{Pr}(C(x; \theta) \neq C(x'; \theta)) \\
    & \approx &  |\B_+^x| - \sum_{x' \in \B_+^x} \mathbf{Pr}(C(x; \theta) = C(x'; \theta)) \\
    & \leq & |\B_+^x| - \sum_{x' \in \B_+^x} \frac{\exp(\phi_{x,x'})}{\sum_{x' \in \B}\exp(\phi(x,x'))}  \\
    \implies R(\theta) & \equiv & \, \max_{\theta} \mathbb{E}_x \left [\sum_{x' \in \B_+^x} \frac{\exp(\phi_{x,x'})}{\sum_{x' \in \B}\exp(\phi(x,x'))} \right] 
\end{IEEEeqnarray*}
where we used the softmax similarity between samples $x,x'$ in the feature space as a proxy for the equality of their classifier outputs and changed $\max$ to sum with the bound. 
Under these specific assumptions, we can now see that the input-regularization objective $R(\theta)$ is strongly reminiscent of our sample consistency loss. Using Eq.~\eqref{eq:sc_loss}, we minimize the negative log-likelihood of the similarity probability, which is equivalent to maximizing the similarity probability of like samples. Therefore, our sample consistency objective is akin to minimizing an upper bound on the input consistency regularization proposed in \cite{wei2020theoretical}.
{Furthermore, optimizing such an objective is shown to achieve bounded target error for unsupervised domain adaptation. Specifically, under the assumption that the psuedo label accuracy on target data is above a certain threshold, \cite{wei2020theoretical} showed that bounded error on target data is achievable using the consistency regularization (Theorem 4.3 ). In \Ours{}, we realize this assumption by first training the feature extractor only using supervised ($\LL_{sup}$) and adversarial ($\LL_{adv}$) losses as explained in \secref{subsec:memory} before introducing our proposed sample consistency loss.
}
To the best of our knowledge, we are the first to instantiate the regularization proposed in \cite{wei2020theoretical} for large scale domain adaptation, and showcase its effectiveness in achieving significant empirical gains.



\section{Experiments and Analysis}
\label{sec:expmnts}

\shortpara{Datasets} Consistent with the key motivations that distinguish \Ours{} from prior literature in domain adaptation, we focus on large-scale datasets with many categories to underline its benefits. 

{\em \dnet} \cite{peng2019moment} is a large-scale dataset for UDA covering 6 domains and a total of 500k images from 345 different categories. It is an order of magnitude larger compared to prior benchmarks and serves as a useful testbed for evaluating many-class adaptation models. We follow the protocol established in prior works \cite{saito2019semi, prabhu2021sentry, tan2020class} to use data from 4 domains, namely real (\textbf{R}), clipart (\textbf{C}), sketch (\textbf{S}) and painting (\textbf{P}), showing results on all 12 transfer tasks across these 4 domains. In \tabref{tab:domainnet126} in the {supplementary material}, we also provide results using a 126-class subset of \dnet{} which contains much lesser label noise~\cite{saito2019semi, yang2020mico, liang2020combating}. Nevertheless, our benefits persist on both these splits.

{\em CUB (Caltech-UCSD birds)} \cite{WahCUB_200_2011} is a challenging dataset originally proposed for fine-grained classification of 200 categories of birds, while {\em CUB-Drawings} \cite{PAN} consists of paintings corresponding to the 200 categories of birds in CUB. We use this dataset pair, consisting of 14k images in total, for evaluation of adaptation on images with fine-grained categories. This setting can be challenging as appearance variations across species can be subtle, while pose variations within a class can be high. Thus, discriminative transfer requires precisely mapping category-specific information from source to target to avoid negative transfer. Results on other fine-grained datasets like Birds-123 and CompCars~\cite{yang2015large} are present in \secref{sec:compcars} in the supplementary material.

\vspace{2pt}
\shortpara{Training Details} We use a Resnet-50~\cite{he2016deep} backbone pretrained on Imagenet, followed by a projection layer as the encoder $\E$ to obtain 256 dimensional feature embeddings. The discriminator $\G$ is implemented using an MLP with two hidden layers of 1024 dimensions. We use a standard batch size of 32 for both source and target in all experiments and for all methods. The reported accuracies are computed on the complete unlabeled target data for CUB-200 dataset following established protocol for UDA \cite{CDAN,xu2019larger,PAN,saito2018maximum}, and the provided testset for \dnet{}. The crucial hyper-parameters in our method are $\lambda_{sc}$, temperature $\tau$ and memory bank size $|\M|$. For all datasets, we choose $\lambda_{sc}=0.1$ and $\tau=0.07$ based on the adaptation performance on the $C \rightarrow D$ setting on the CUB-200 dataset. 
We use a memory bank size of 48k on \dnet{} dataset, but 24k on CUB-200 dataset owing to its smaller size. For experiments on \Ours{}, we report mean and standard deviation over 3 random seeds. We compare \Ours{} against traditional adversarial approaches ({DANN}~\cite{DANN}, {CDAN}~\cite{CDAN}, {MCD}~\cite{saito2018maximum}) as well as the current state-of-the art ({SAFN}~\cite{xu2019larger}, {BSP}~\cite{chen2019transferability}, {RSDA}~\cite{gu2020spherical}, CAN~\cite{kang2019contrastive}, ILADA~\cite{ILA}, {FixBi}~\cite{na2021fixbi}, HDAN~\cite{cui2020hda} and ToAlign~\cite{wei2021toalign}). 
%


\begin{table}[!tbp]
    \centering
    \captionsetup{width=\textwidth, font=footnotesize}
    \caption{Accuracy scores on \dnet-345 using Resnet-50 backbone. Best values are in \textbf{bold} and the next best are \underline{underlined}. \Ours{} performs better than all other methods on most of the tasks. $^\dagger{}$Uses hierarchical label annotation. $^\ddagger$prediction uses ensemble classifiers. $^\mathsection$Uses class-balanced sampling.}
    \label{tab:domainnet345}
    \resizebox{\textwidth}{!}{
    \begin{tabular}{@{} l *{17}{c} @{} }
        \toprule
        Source & \multicolumn{3}{c}{\textbf{Real$\rightarrow$}} && \multicolumn{3}{c}{\textbf{Clipart$\rightarrow$}} && \multicolumn{3}{c}{\textbf{Painting$\rightarrow$}} && \multicolumn{3}{c}{\textbf{Sketches}$\rightarrow$} && \\
        \cline{2-4} \cline{6-8} \cline{10-12} \cline{14-16}
        Target & {C} & {P} & {S} && {R} & {P} & {S} && {R} & {C} & {S} && {R} & {C} & {P} && Avg. \\
        \midrule
        ResNet-50& 41.61& 42.79& 29.66& & 42.41& 27.24& 32.15& & 49.52& 32.55& 26.73& & 38.75& 40.89& 27.5& & 35.98 \\
        MSTN~\cite{xie2018learning} & 27.25& 32.98& 24.35& & 28.17& 21.14& 24.15& & 30.74& 19.85& 22.5& & 24.31& 26.22& 23.56& & 25.44 \\
        RSDA~\cite{gu2020spherical} & 27.28 & 35.83& 24.35& & 36.98& 24.94& 31.12& & 41.32& 26.1& 24.71& & 29.46& 26.22& 27.79& & 29.68 \\
        BSP~\cite{chen2019transferability} & 34.51 & 39.14& 27.57& & 40.56& 26.71& 30.72& & 40.83& 24.56& 26.85& & 36.54& 32.37& 28.08 & & 32.37 \\
        MCD~\cite{saito2018maximum}$^\ddagger$ & 36.34& 36.58& 24.95& & 40.32& 25.83& 32.12& & 43.65& 29.66& 25.7& & 34.16& 39.11& 26.89& & 32.94$^\ddagger$ \\
        ILADA~\cite{ILA}$^\mathsection$ & 46.45 & 39.01 & 35.4&& 47.94 &26.68 &36.33 &&43.00 & 26.62 & 27.3 && 48.85 & 47.68 & 32.23 && 38.12$^\mathsection$  \\
        SAFN~\cite{xu2019larger} & 38.11& 45.96& 29.20 & & 45.96& 30.00 & 34.65& & {54.44}& 34.74& 30.64& & 45.29& 47.43& 38.01& & 39.54 \\
        DANN~\cite{DANN} & 45.93& 44.51& 35.47& & 46.85& 30.52& 36.77& & 48.02& 34.76& 32.15& & 47.1& 46.45& 38.47& & 40.58 \\
        CAN~\cite{kang2019contrastive}$^\mathsection$ & 40.71 &37.77 &33.7 && \textbf{54.93} &31.41 &37.37 && 51.05 &33.64 &30.95 && \underline{52.13} &42.19 &32.04 && 39.82$^\mathsection$  \\
        PAN~\cite{PAN}$^\dagger$ & 49.25 &48.18 &36.46 && 49.66 &33.27 &38.78 && 51.89 &36.01 &32.94 && 49.12 &50.94 &39.89 && 43.03 $^\dagger$ \\
        CDAN~\cite{CDAN} & 50.15& 48.35& 39.01& & 50.02& 33.39& 39.3& & 52.21& 36.44& 33.68& & 48.46& 49.27& 38.65& & 43.24 \\
        HDAN~\cite{cui2020hda} & 46.30 & 47.52 & 34.39 & & 49.91 & 33.98 & 37.98 & & \underline{55.26} & {40.82} & 32.77 & & 49.04 & 49.77 & 40.04 & & 43.15 \\
        FixBi~\cite{na2021fixbi}$^\ddagger$ & \underline{51.18}& {49.19}& \underline{39.65}& & 50.02& \underline{34.59}& {41.17}& & 52.21& 36.44& {33.68}& & 50.84& {53.51}& \underline{41.67}& & {44.51}$^\ddagger$ \\
        ToAlign~\cite{wei2021toalign} & {50.82}& \underline{50.72}& {35.17}& & 49.52 & {33.88}& \underline{41.41}& & \textbf{57.92} & \textbf{43.51} & \underline{36.29}& & {47.96} & \textbf{55.46}& {41.61}& & \underline{45.45} \\
        \midrule
        MemSAC [Ours] & \textbf{54.34}$^{\pm.5}$& \textbf{52.27}$^{\pm.3}$& \textbf{41.74}$^{\pm.3}$& & \underline{54.4}$^{\pm.3}$& \textbf{36.87}$^{\pm.4}$& \textbf{42.45}$^{\pm.0}$& & 53.24$^{\pm.2}$& \underline{41.39}$^{\pm.4}$& \textbf{37.22}$^{\pm.2}$& & \textbf{53.33}$^{\pm.3}$& \underline{55.31}$^{\pm.2}$& \textbf{44.56}$^{\pm.3}$& & \textbf{47.26} \\
        \midrule 
        \textcolor{lightgray}{Tgt. Supervised} & \textcolor{lightgray}{72.59} & \textcolor{lightgray}{62.66} &\textcolor{lightgray}{65.12} & & \textcolor{lightgray}{80.92} & \textcolor{lightgray}{62.66} & \textcolor{lightgray}{65.12} & & \textcolor{lightgray}{80.92} & \textcolor{lightgray}{72.59} & \textcolor{lightgray}{65.12} & & \textcolor{lightgray}{80.92} & \textcolor{lightgray}{72.59} & \textcolor{lightgray}{62.66 } && \textcolor{lightgray}{70.32} \\
        \bottomrule
    \end{tabular}
    }
    \vspace{-12pt}
\end{table}
\vspace{2pt}
\noindent {\bf MemSAC significantly outperforms others on many-class adaptation} The results for the 12 transfer tasks on \dnet{} are provided in \tabref{tab:domainnet345}. 
Firstly, methods such as RSDA (29.68\%) and SAFN (39.54\%) that achieve best performance on smaller scale datasets (like Office-31~\cite{saenko2010adapting} and visDA-2017~\cite{peng2017visda}) provide only marginal or no benefits over the more traditional adversarial approaches such as DANN (40.58\%) and CDAN (43.24\%) on \dnet{} with 345 classes, indicating that large-scale datasets need different techniques for adaptation. Next, we compare against PAN~\cite{PAN}, which requires a label hierarchy as additional information for training. For this supervision, we use the one level of hierarchy proposed in \dnet{}~\cite{peng2019moment}. Even when provided with access to hierarchical grouping labels in source, PAN (43.03\%) achieves no improvement over CDAN (43.24\%). In contrast, our method \Ours{} that combines global adaptation using a conditional adversarial approach and local alignment using sample consistency to alleviate negative achieves an average accuracy of 47.26\%, with a significantly better performance than all the prior approaches across most of the tasks. 

\vspace{2pt}
\noindent {\bf MemSAC achieves new state-of-the-art in fine-grained adaptation} We also illustrate the benefit of using \Ours{} for adaptation on fine-grained categories in \tabref{tab:cub200_result} on the CUB-Drawings dataset. Although fine-grained visual recognition is a well-studied area \cite{zhang2012pose,zhang2014part,branson2014bird,chen2019destruction,dubey2018maximum}, domain adaptation for fine grained categories is a relevant but less-addressed problem. Notably, methods like MCD, SAFN and RSDA perform worse or only marginally better than a source only baseline. PAN~\cite{PAN} uses supervised hierarchical label relations in source across 3 levels and obtains an average accuracy of 62.96\%, while \Ours{} obtains a state-of-the art accuracy of 67.95\% using only single level source labels, thus outperforming all prior approaches on this challenging setting with minimal assumptions. This underlines the benefit of enforcing sample consistency using \Ours{} for adaptation even in the presence of fine-grained categories in order to effectively counter negative alignment issues. 
%

\begin{table}[!t]
  \centering
   \captionsetup{width=\textwidth, font=footnotesize}
  \caption{Results on fine-grained adaptation on 200 categories from CUB-Drawings dataset. Bold and underline indicate the best and second best methods respectively. $\dagger$Uses hierarchical label annotation. $^\mathsection$Uses class-balanced sampling.}
  \label{tab:cub200_result}
  \resizebox{\textwidth}{!}{
  \begin{tabular}{@{} l *{30}{c} @{}} 
    \toprule
    & Resnet-50 && MCD&& SAFN&& CAN$^\mathsection$&& RSDA&& DANN&& HDAN&& FixBi&& CDAN&& ToAlign&& PAN$^\dagger$&& \Ours{} \\
     & && \cite{saito2018maximum} && \cite{xu2019larger} && \cite{kang2019contrastive} && \cite{gu2020spherical} && \cite{DANN} && \cite{cui2020hda} && \cite{na2021fixbi} && \cite{CDAN} && \cite{wei2021toalign} && \cite{PAN} && \\
    \midrule
    C $\rightarrow$ D & 60.88 && 50.18 && 60.29 && 52.18 && 61.04 && 62.09 && 60.25 && 68.20 && 68.12 && 64.43 && \underline{70.53} && \textbf{73.97} \\
    D $\rightarrow$ C & 42.07 && 38.56 && 41.34 && 50.05 && 44.20 && 47.73 && 52.40 && 49.47 && 53.83 && 50.54 && \underline{55.38} && \textbf{61.94} \\
    Avg. & 51.47 && 44.37 && 50.82 && 51.11 && 52.62 && 54.91 && 56.33 && 58.84 && 60.98 && 57.48 && \underline{62.96} && \textbf{67.95} \\
    \bottomrule
  \end{tabular}}
  \vspace{-11pt}
\end{table}

\vspace{2pt}
\noindent {\bf MemSAC complements multiple adaptation methods} The proposed memory-augmented consistency loss is generic enough to improve many adaptation backbones. As shown in \tabref{tab:dnet+dann} for the case of R$\rightarrow$C and C$\rightarrow$R transfer tasks from \dnet{}, \Ours{} can be used with most adversarial as well as MMD based approaches. \Ours{} improves adversarial approaches DANN and CDAN by 3.35\% and 4.29\% respectively, and MMD-based approach CAN by 1.75\% indicating that our proposed framework is competitive yet complementary to many existing adaptation approaches.
Complete table for all the 12 transfer tasks is provided in \tabref{tab:net+dann_sup} in the {supplementary material}. 

\vspace{2pt}
\noindent {\bf MemSAC improves adaptation even with larger backbones} We employ Resnet-101 as a backbone in \tabref{tab:dnet+dann} and compare against other adaptation approaches with the same backbone. We note that the benefits obtained by \Ours{} over prior adaptation approaches 
also hold for larger backbones, as shown for R$\rightarrow$C and C$\rightarrow$R of \dnet{} dataset, and complete table containing results on all transfer tasks is presented in \tabref{tab:resnet101_supp} in the {supplementary material}.

\begin{table*}[tbp]
    \begin{center} 
    \captionsetup{width=\textwidth, font=footnotesize}
    \caption{{MemSAC is also complementary to most adversarial and adaptation methods, as shown in (\subref{tab:dnet+dann}). We show the results using a larger backbone (Resnet-101) for training in (\subref{tab:resnet101}). \Ours{} adds negligible memory and time overhead to the training even with large queues, and zero overhead during inference, as shown in (\subref{tab:inftime}})
    }
    \begin{minipage}[t]{0.37\hsize}\centering
    \captionsetup{width=0.9\textwidth, font=footnotesize}
    \subcaption{MemSAC complements existing UDA methods.}
    \label{tab:dnet+dann}
    \resizebox{0.9\textwidth}{!}{
    \begin{tabular}{@{} l *{17}{c} @{} }
        \toprule
        & R$\rightarrow$C && C$\rightarrow$R && Avg.  \\
        \midrule
        DANN~\cite{DANN} & 45.93 && 46.85 && 46.39 \\
        DANN+\Ours{} & 49.67 && 49.81 && \textbf{49.74}(\textcolor{blue}{+3.35\%}) \\
        \midrule
        CAN~\cite{kang2019contrastive} & 40.71 && {54.93} && 47.82 \\
        CAN+\Ours{} & 43.79 && 55.36 && \textbf{49.57}(\textcolor{blue}{+1.75\%}) \\
        \midrule
        CDAN~\cite{CDAN} & 50.15 && 50.02 && 50.08 \\
        CDAN+\Ours{} & 54.34 && 54.40 && \textbf{54.37}(\textcolor{blue}{+4.29\%}) \\
        \bottomrule
        \end{tabular}}
    \end{minipage}
    \hfill
    \begin{minipage}[t]{0.27\hsize}
    \centering
    \captionsetup{width=.8\textwidth, font=footnotesize}
    \subcaption{Results using Resnet-101 backbone}
    \label{tab:resnet101}
    \resizebox{0.9\textwidth}{!}{
    \begin{tabular}{@{} l *{17}{c} @{} }
        \toprule
        & R$\rightarrow$C && C$\rightarrow$R && Avg.  \\
        \midrule
        Resnet-101 & 45.62 && 41.96 && 43.79 \\
        DANN~\cite{DANN} & 47.71 && 48.33 && 48.02  \\
        MCD~\cite{saito2018maximum} & 41.11 && 40.77 && 40.94 \\
        CDAN~\cite{CDAN} & 52.47 && 46.63 && 49.55 \\
        SAFN~\cite{xu2019larger} & 44.93 && 37.20 && 41.06 \\
        ToAlign\cite{wei2021toalign} & 50.09 && 50.23 && 50.16 \\
        \Ours{} & \textbf{56.25}  && \textbf{53.52} && \textbf{54.88} \\
        \bottomrule
        \end{tabular}}
    \end{minipage}
    \hfill
    \begin{minipage}[t]{0.34\hsize}\centering
        \captionsetup{width=0.9\textwidth, font=footnotesize}
        \subcaption{Training times of various methods.} 
        \label{tab:inftime}
        \vspace{8pt}
        \resizebox{\textwidth}{!}{
            \begin{tabular}{@{} l  c l c @{} }
                \toprule
                Method & Peak GPU Mem. & Training Time & Avg. Acc\\
                \midrule
                DANN & 7.9GB & 11.7 Hrs & 40.58\% \\
                CDAN & 8.2GB & 12 Hrs & 43.24\% \\
                PAN  & 8.9GB & 16.2 Hrs & 43.03\% \\
                ToAlign & 9.22GB &  24.21 Hrs & 45.45\% \\
                \Ours{} & 8.5GB & 12.63 Hrs & 47.26\% \\
                \bottomrule
            \end{tabular}
            }
    \end{minipage}
    \end{center}
    \label{tab:domainnet_ablation}
    \vspace{-20pt}
\end{table*}

\subsection{Analysis and Discussion} 
\label{subsec:analysis}

\shortpara{Ablation studies} We show the influence of various design choices of our method in \tabref{tab:ablations} on the CUB-200 dataset. First, we show in \tabref{tab:lossComponent} that both the global domain adversarial method, which we implement using CDAN, as well as local sample level consistency loss are important to achieve best accuracy, as evident from the drop in accuracy without either of those components. Next, we investigate the effect of the temperature parameter $\tau$ in \tabref{tab:tau} which we use to suitably scale the contributions of positive and negative pairs in $\LL_{sc}$ loss function (Eq.~\eqref{eq:sc_loss}). We find that $\tau=0.07$ gives the best performance on the cosine similarity metric. 
%
Finally, in \tabref{tab:similarity}, we note that the performance using other choices of the similarity function $\phi(.)$, namely \textit{Euclidean} similarity and \textit{Gaussian} similarity is inferior to using \textit{Cosine} similarity. 
We also observed that \textit{cosine} similarity is more stable to train under severe domain shifts. 

\begin{figure*}[!t]
     \centering
    \begin{minipage}[b]{0.3\textwidth}
        \centering
        \includegraphics[width=\textwidth]{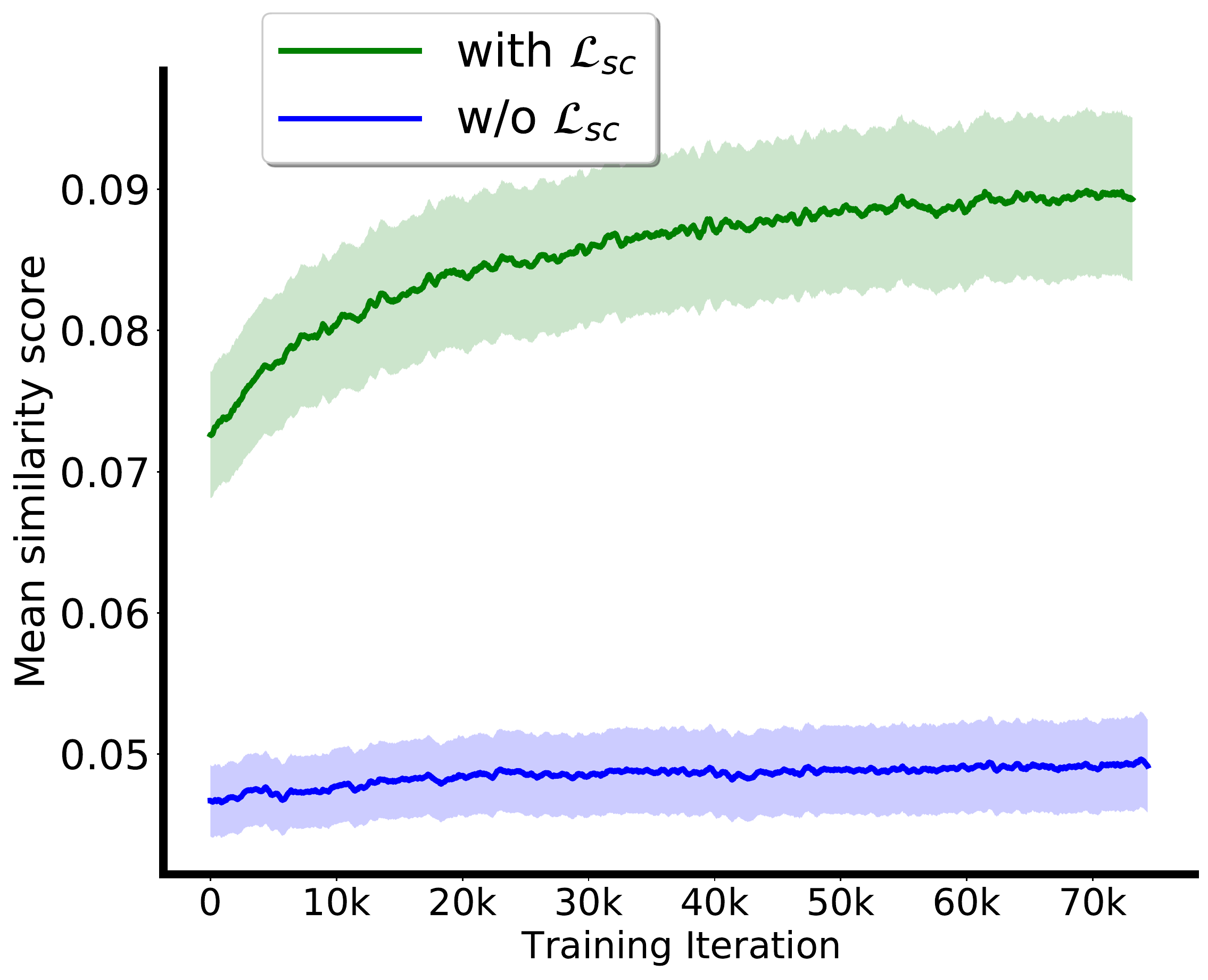}
        \captionsetup{width=1\textwidth, font=footnotesize}
        \caption{Mean similarity score for \emph{within-class} samples vs. training iteration shown for \textbf{D}$\rightarrow$\textbf{C} on CUB-Drawings.
        }
        \label{fig:simScores}
    \end{minipage}
    \hfill
    \begin{minipage}[b]{0.3\textwidth}
        \centering
        \includegraphics[width=\textwidth]{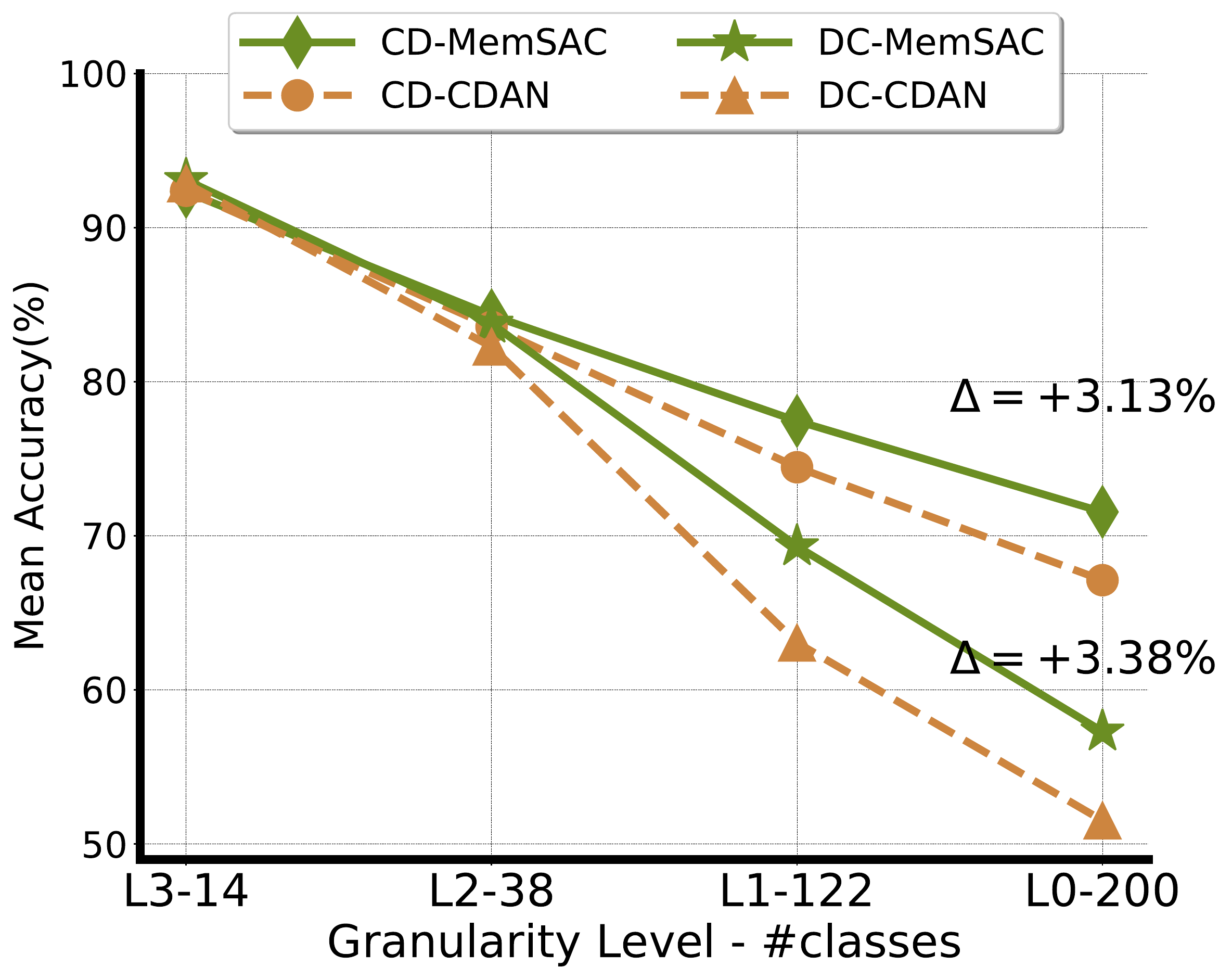}
        \captionsetup{width=0.95\textwidth, font=footnotesize}
        \caption{Comparison of accuracy vs. granularity of labels on CUB-Drawings dataset for 4 levels of label hierarchy. }
        \label{fig:hierarchyAcc}
     \end{minipage}
     \hfill
     \begin{minipage}[b]{0.37\textwidth}
        \centering
        \includegraphics[width=\textwidth]{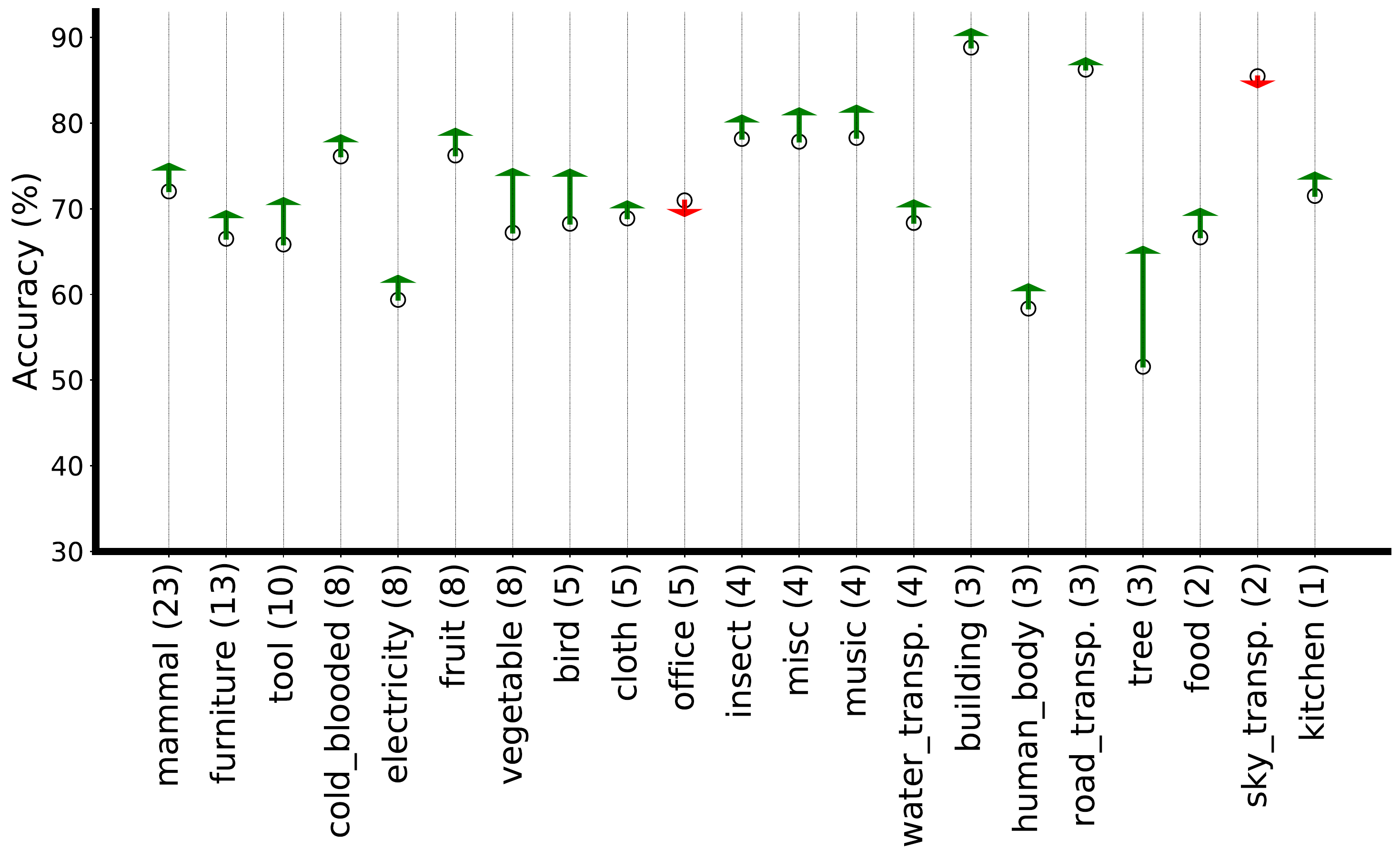}
        \captionsetup{width=0.95\textwidth, font=footnotesize}
        \caption{Category wise gain/drop in accuracy on \textbf{R}$\rightarrow$\textbf{C} on \dnet{}, compared to CDAN~\cite{CDAN}.}
        \label{fig:categoryAcc}
     \end{minipage}
\vspace{-12pt}
\end{figure*}

\vspace{2pt}
\noindent {\bf Why does \Ours{} help with large number of classes?}
We propose our sample consistency loss in \eqref{eq:sc_loss} to encourage tighter clustering of samples within each class, which is important in many-class datasets where class confusion is a significant problem.  
The main motivation of the proposed sample consistency loss is to bring {within-class} samples (that is, samples from the same class across source and target domains) closer to each other, so that a source classifier can be transferred to the target. 
To understand this further, in \figref{fig:simScores}, we plot the \emph{mean similarity score} during the training process. We define the \emph{mean similarity score} as $\sum_{i \in \M_{+}^j} \phi_{ij}$, averaged over all the target samples $j \in \B_t$ in a mini-batch, which indicates the affinity score between same-class samples across domains. We observe that using the proposed loss, the similarity score is much higher and improves with training compared to the baseline without the consistency loss, which reflects in the overall accuracy (\tabref{tab:domainnet345}, \tabref{tab:cub200_result}).

\vspace{2pt}
\shortpara{\Ours{} achieves larger gains with finer-grained classes} We show the appreciating benefits provided by \Ours{} as the fine-grainedness of the dataset becomes more pronounced. For this purpose, we chose the 4 levels of label hierarchy provided by PAN~\cite{PAN} on the CUB-Drawings dataset. The levels L3, L2, L1 and L0 contain different granularity of bird species, grouped into 14, 38, 122 and 200 classes, respectively. The L0 level contains the finest separation of classes, while the level L3 with 14 classes contains the coarsest separation.
We observe from \figref{fig:hierarchyAcc} that with coarser granularity, \Ours{} performs as good as the baseline method CDAN, whereas with finer separation of the categories (L3 $\rightarrow$ L0), use of sample consistency loss provides much higher benefit ($>3\%$ improvement on both tasks). This confirms our intuition that sample level consistency benefits accuracies in fine-grained domain adaptation.

\begin{table*}[!tbp]
\centering
    \captionsetup{width=\textwidth, font=footnotesize}
    \caption{{\bf Ablation results}. Effect of (\subref{tab:lossComponent}) Loss coefficients, (\subref{tab:tau}) temperature scaling, and (\subref{tab:similarity}) choice of similarity functions on accuracy of \Ours{} on the CUB-Drawing adaptation. }
    \label{tab:allablations}
    \begin{minipage}[b]{0.32\hsize}\centering
        \captionsetup{width=\textwidth, font=footnotesize}
        \subcaption{Effect of various components of loss function in \eqref{eq:total_loss}.}
        \label{tab:lossComponent}
        \resizebox{\textwidth}{!}{
        \begin{tabular}{@{} l *{5}{c} @{} }
            \toprule
            Method & $\LL_{adv}$ & $\LL_{sc}$ & C$\rightarrow$D & D$\rightarrow$C & Avg. Acc\\
            \midrule
            Source & \cross & \cross & 60.88 & 42.07 & 51.47 \\
            CDAN   & \tick  & \cross & 68.12 & 53.83 & 60.98 \\
            $\LL_{sc}$ Only     & \cross & \tick  & 64.45 & 41.13 & 52.79 \\
            \Ours{}      & \tick  & \tick  & \textbf{73.97} & \textbf{61.94} & \textbf{67.95} \\
            \bottomrule
        \end{tabular}
        }
    \end{minipage}
    \hfill
    \begin{minipage}[b]{0.28\hsize}\centering
        \captionsetup{width=0.95\textwidth, font=footnotesize}
        \subcaption{Effect of the temperature $\tau$ in \eqref{eq:sc_loss}.}
        \label{tab:tau}
        \resizebox{0.9\textwidth}{!}{
        \begin{tabular}{@{} r *{3}{c} @{} }
            \toprule
            $\tau$ & C$\rightarrow$D & D$\rightarrow$C & Avg. Acc\\
            \midrule
            $1.0$ & 68.36 &	53.46 &	60.91 \\
            $0.07$ & \textbf{73.97} & \textbf{61.94} & \textbf{67.95} \\
            $0.007$ & 71.25 & 57.21 &  64.23 \\
            \bottomrule
        \end{tabular}
        }
    \end{minipage}
    \hfill
    \begin{minipage}[b]{0.36\hsize}\centering
    \captionsetup{width=0.95\textwidth, font=footnotesize}
        \subcaption{Accuracy using various choices for $\phi_{ij}$.}
        \label{tab:similarity}
        \resizebox{\textwidth}{!}{
        \begin{tabular}{@{} l *{4}{c} @{} }
            \toprule
            Similarity & $\phi_{ij}$ & C$\rightarrow$D & D$\rightarrow$C & Avg. Acc\\
            \midrule
            Inv. Euc. & $(1 + ||f_i-f_j||^2)^{-1}$ & 71.00 & 57.21 & 64.23 \\
            \\
            Gaussian & $exp(-||f_i - f_j||^2)$ & 70.10 & 50.84 & 60.47 \\
            \\
            Cosine & $f_i \cdot f_j$ & \textbf{73.97} & \textbf{61.94} & \textbf{67.95} \\
            \bottomrule
        \end{tabular}
        }
    \end{minipage}
\label{tab:ablations}
\vspace{-4pt}
\end{table*}

\vspace{2pt}
\noindent {\bf \Ours{} alleviates class confusion for similar classes} In \figref{fig:categoryAcc} we use the \dnet{} dataset to show the accuracies on every \textit{coarse} category, along with the number of finer classes in each coarse category. 
We find that \Ours{} provides consistent improvement over CDAN (marked by \textcolor{green}{$\uparrow$}) on most categories and any drops in accuracy (marked by \textcolor{red}{$\downarrow$}) are negligible. Our improvements are especially greater on categories with fine-grained classes like \textit{trees (+13.3\%)}, \textit{vegetables (+6.7\%)} and \textit{birds (+5.6\%)}, underlining the advantage of \Ours{} to overcome class confusion within dense categories.

\begin{figure*}[!t]
     \centering
     \begin{minipage}[b]{0.4\textwidth}
        \centering
        \includegraphics[width=\textwidth]{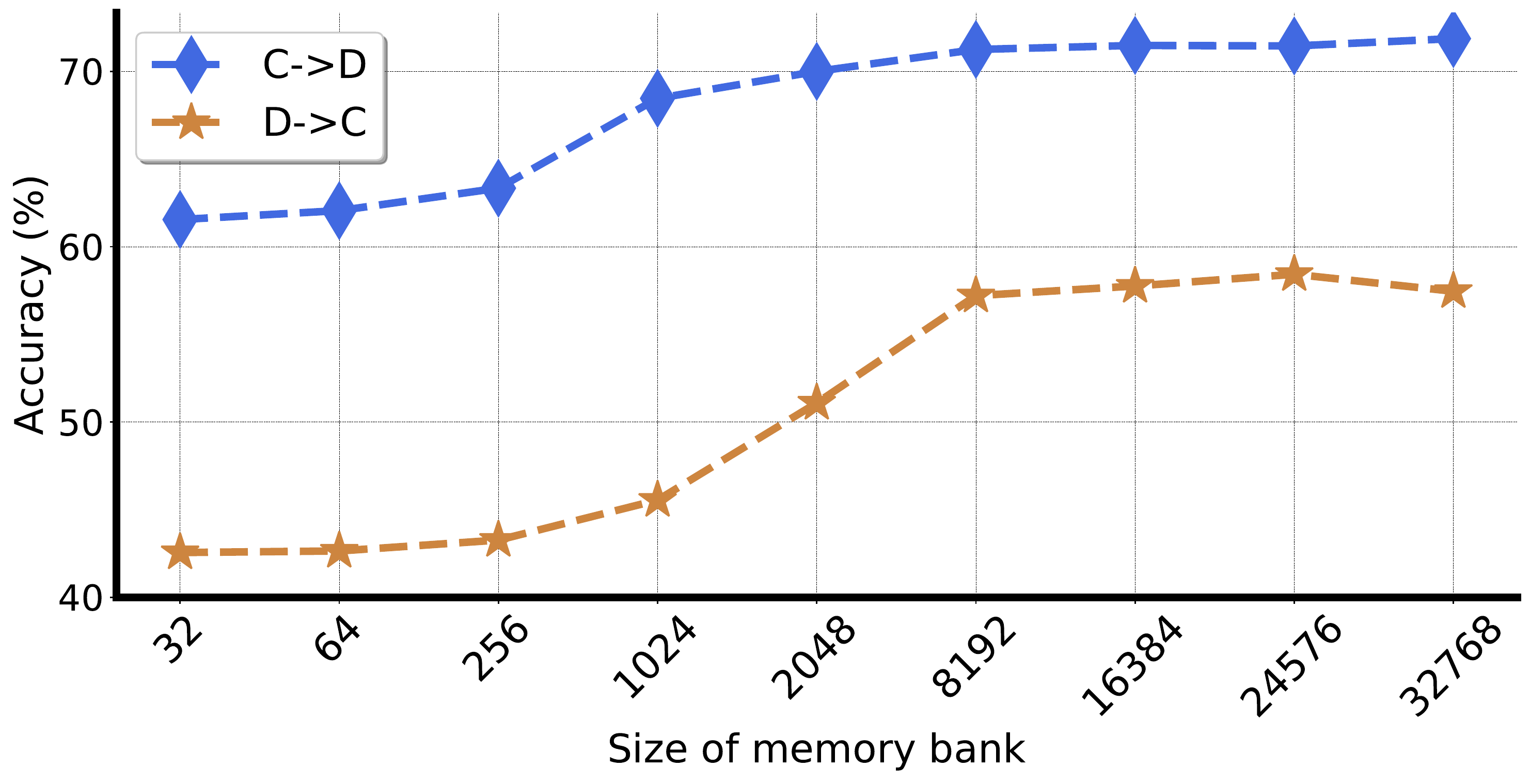}
        \captionsetup{width=\textwidth, font=footnotesize}
        \caption{Effect of memory bank size on CUB-Drawings dataset.}
        \label{fig:membankSize}
     \end{minipage}
     \hfill
     \begin{minipage}[b]{0.55\textwidth}
     \centering
     \begin{subfigure}[t]{0.32\textwidth}
        \centering
        \includegraphics[width=1\textwidth]{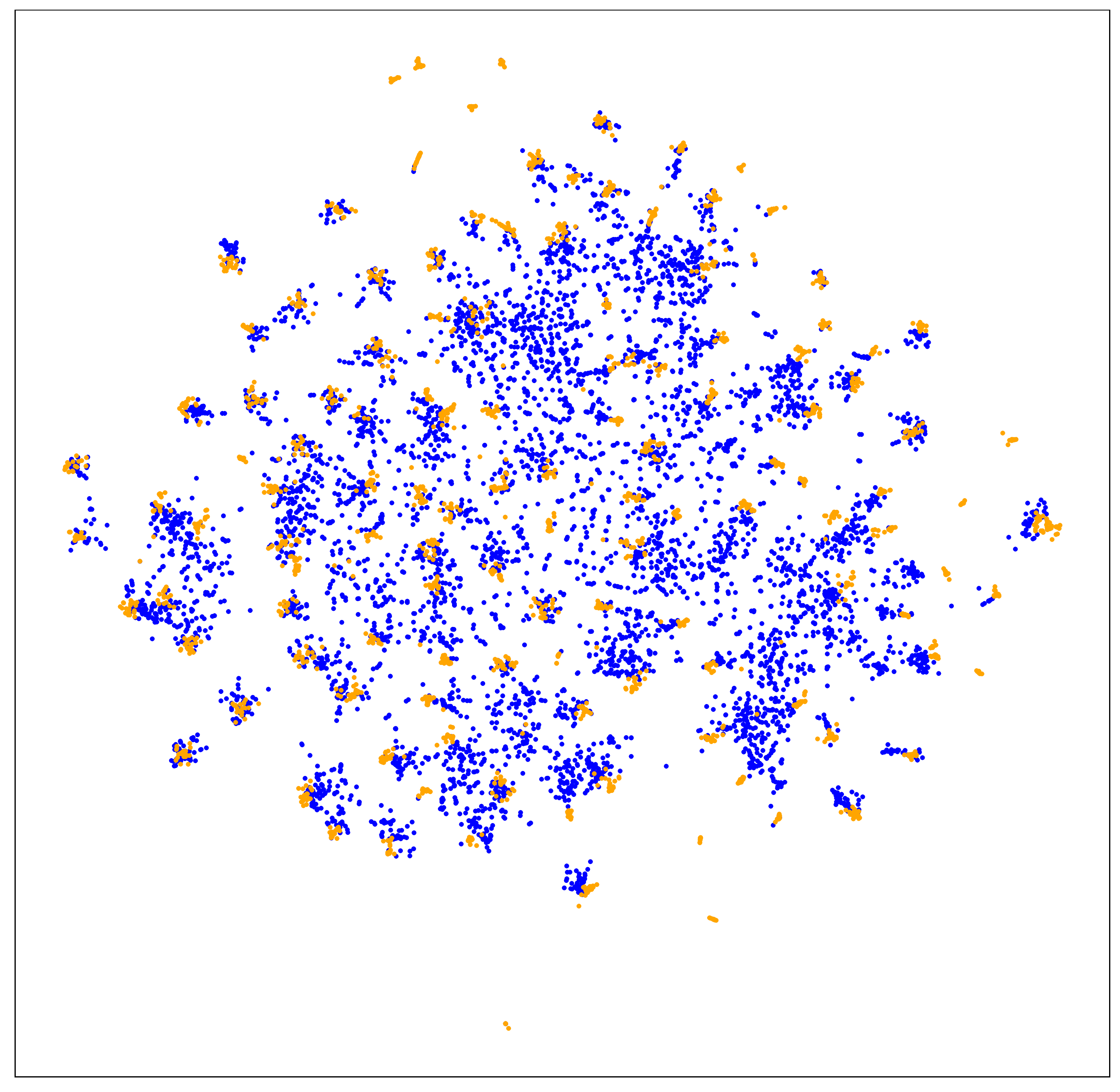}
        \captionsetup{width=\textwidth, font=scriptsize}
        \subcaption{Source only}
        \label{fig:baseline_tsne}
     \end{subfigure}
    \begin{subfigure}[t]{0.32\textwidth}
        \centering
        \includegraphics[width=1\textwidth]{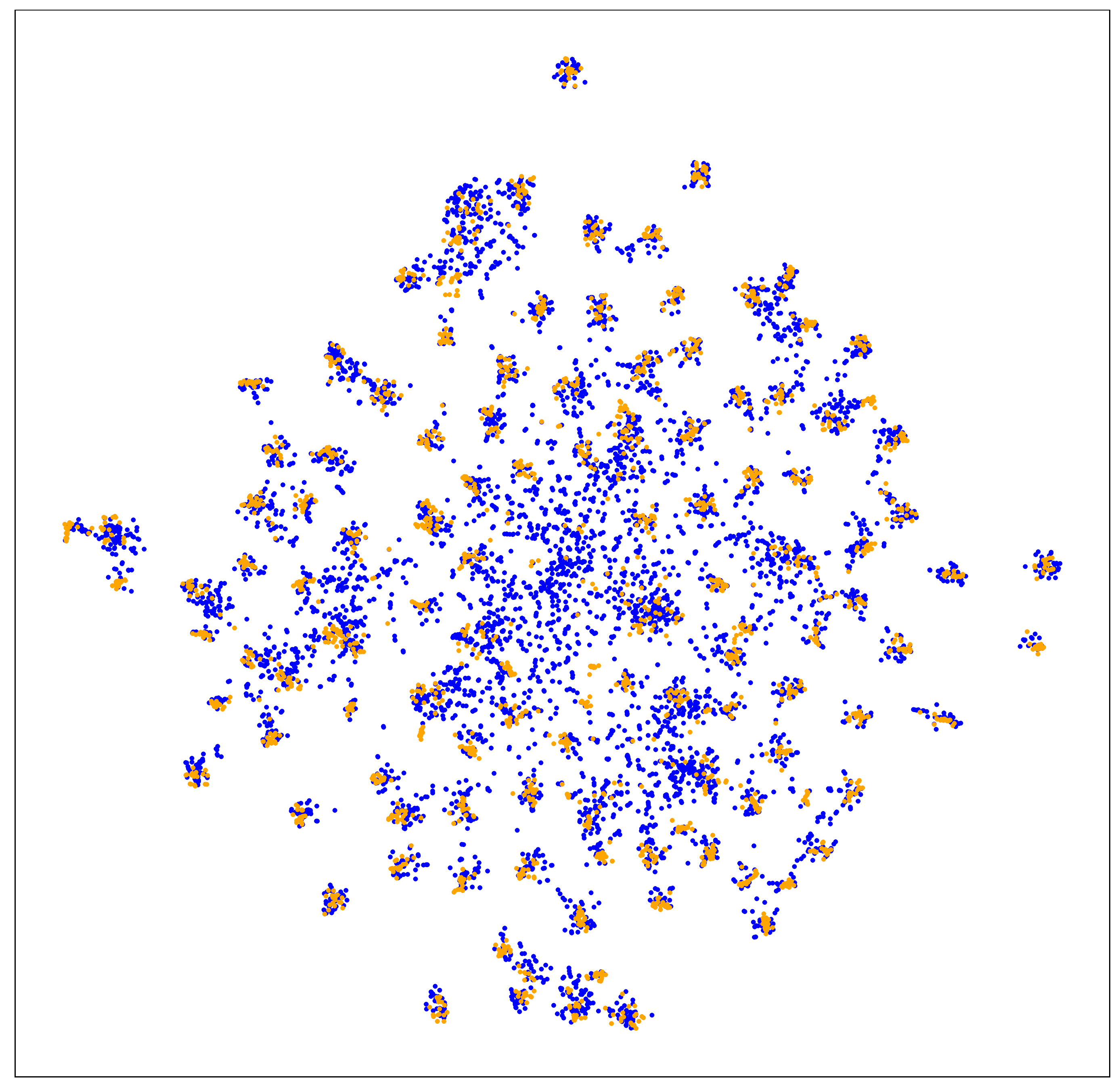}
        \captionsetup{width=\textwidth, font=scriptsize}
        \subcaption{CDAN}
        \label{fig:cdan_tsne}
     \end{subfigure}
     \begin{subfigure}[t]{0.32\textwidth}
        \centering
        \includegraphics[width=1\textwidth]{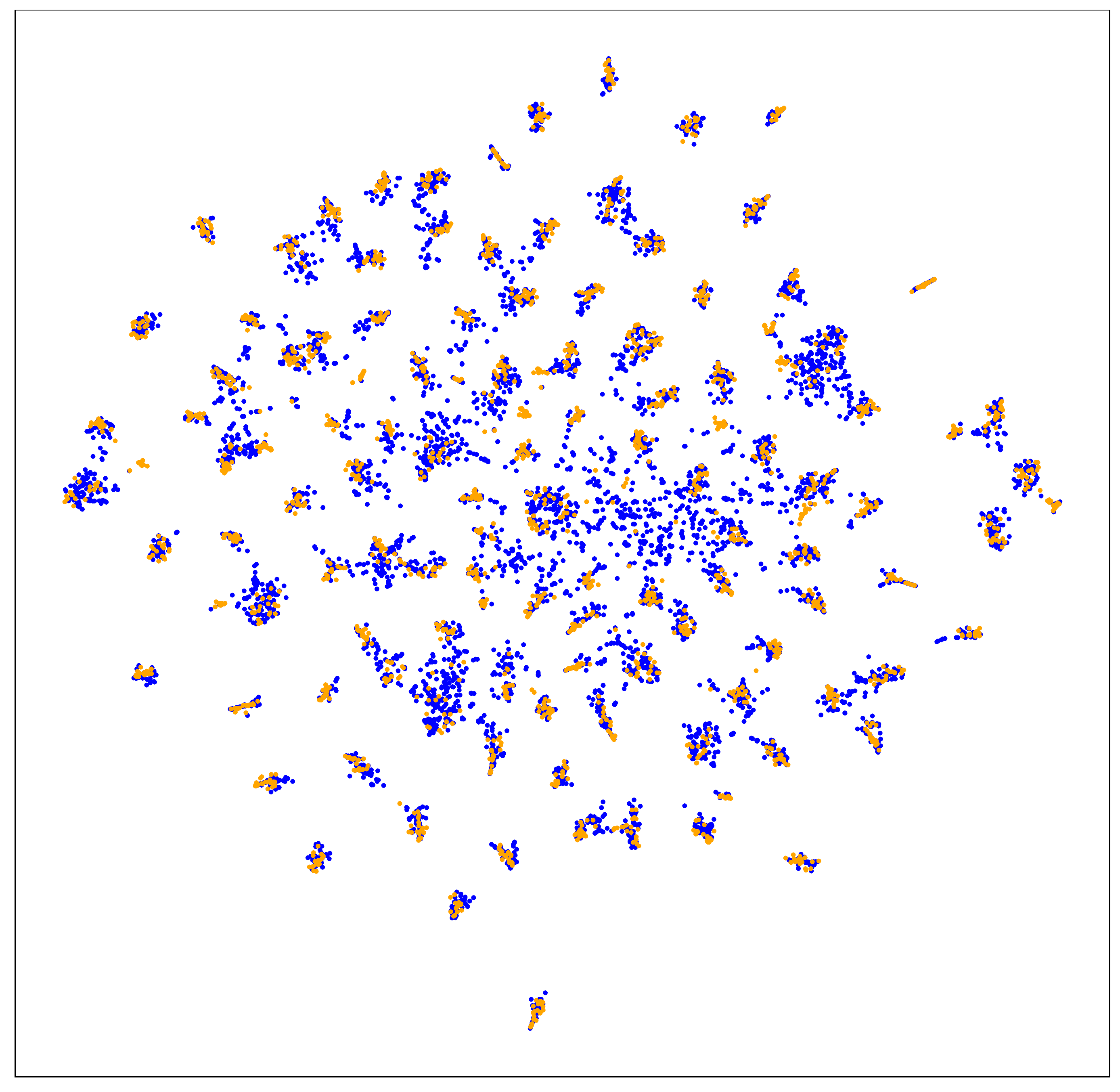}
        \captionsetup{width=\textwidth, font=scriptsize}
        \subcaption{\Ours{}}
        \label{fig:memsac_tsne}
     \end{subfigure}
     \captionsetup{width=\textwidth, font=footnotesize}
     \caption{tSNE for \textbf{R}${\rightarrow}$\textbf{C} on \dnet{}. 
     The two colors are source and target features. Notice improved alignment and feature separation with \Ours{}.}
     \label{fig:tSNE_domainLevel}
    \end{minipage}
\vspace{-0.3cm}
\end{figure*}

\vspace{2pt}
\shortpara{Larger memory banks improve accuracy} A key design choice that we need to make in \Ours{} is the size of the memory bank $\M$. Intuitively, small memory banks would not provide sufficient negative pairs in the sample consistency loss and lead to noisy gradients. We show in \figref{fig:membankSize} for the two tasks in CUB-Drawings that accuracy indeed increases with larger sizes of memory banks (a memory size of 32, which is same as batch-size, indicates no memory at all and performs worse). We also find that the optimum capacity of the memory bank may even be much higher than the size of the dataset. For example, the ``drawing'' domain has around 4k examples, but from \figref{fig:membankSize}, \textbf{D}$\rightarrow$\textbf{C} achieves best accuracy at memory size of 25k.
Since the feature encoder is simultaneously trained while updating memory bank, two copies of the same instance need not necessarily be exact duplicates of each other, but instead provide complementary ``views'' of the same sample. Thus, large queues help in enriching the positive and negative sample set, improving the accuracy.


\vspace{2pt}
\shortpara{Computational cost and resources} We show the training time and GPU memory consumed for \Ours{} compared to other baseline approaches in \tabref{tab:inftime}. In summary, \Ours{} incurs negligible overhead in memory during training and no overhead during inference even for large queues.

\shortpara{Limitations, future work and impact } 
Although we report outstanding performance using \Ours{}, we assume that the list of categories present in the data is known beforehand. Therefore, an avenue of future work is to relax this assumptions and extend \Ours{} to open world adaptation approaches.
While domain adaptation may have the positive impact of equitable performance of machine learning across geographic or social factors, \Ours{} shares with other deep domain adaptation approaches the limitation of lack of explainability, which may have a negative impact on applications where decisions based on domain adaptation have a bearing on safety. We further note that significant room for improvement remains in achieving unsupervised domain adaptation that approach fully supervised performances. 

\vspace{-8pt}
\section{Conclusion}

\vspace{-8pt}
We proposed \Ours{}, a simple and effective approach for unsupervised domain adaptation designed to handle a large number of categories. We propose a sample consistency loss that pulls samples from similar classes across domains closer together, while pushing dissimilar samples further apart. Since minibatch sizes are limited, we devise a novel memory-based mechanism to effectively extract similarity relations for a large number of categories. We provide both theoretical intuition and empirical insights into the effectiveness of \Ours{} for large-scale domain alignment and discriminative transfer. In extensive experiments and analysis across the main paper and supplementary, we showcase the strong improvements achieved by \Ours{} over prior works, setting new state-of-the-arts across challenging many-class adaptation on DomainNet (126 and 345 classes) and fine-grained adaptation on CUB-Drawings (200 classes). 

\noindent {\bf Acknowledgements} We thank NSF CAREER 1751365, NSF Chase-CI 1730158, Google Award for Inclusion Research and IPE PhD Fellowship.

%
%
\bibliographystyle{splncs04}
\bibliography{main}

\appendix
\section*{Supplementary}
\begin{appendix}

\renewcommand\thefigure{S.\arabic{figure}}    
\setcounter{figure}{0}    
\renewcommand\thetable{S.\arabic{table}}    
\setcounter{table}{0}

\section{Results on \dnet{}-126}

In the main paper, we choose the complete 345-class split of \dnet{} to report the results. 
In \tabref{tab:domainnet126}, we show that the benefits using \Ours{} persist even on the split with 126 classes which has much lesser label noise, as proposed in prior works like \cite{saito2019semi, yang2020mico, liang2020combating}. For this experiment, we choose the recommended train-test splits in the official \dnet{} website for the domains real (\textbf{R}), clipart (\textbf{C}), sketch (\textbf{S}) and painting (\textbf{P}). \Ours{} achieves an accuracy of 64.76\% classes, which is $3\%$ larger than the next best approach, PAN (61.75\%). 
\begin{table*}[!h]
    \begin{center} 
    \setlength{\tabcolsep}{3pt}
    \resizebox{\textwidth}{!}{
    \begin{tabular}{@{} l *{17}{c} @{} }
        \toprule
        Source & \multicolumn{3}{c}{\textbf{Real$\rightarrow$}} && \multicolumn{3}{c}{\textbf{Clipart$\rightarrow$}} && \multicolumn{3}{c}{\textbf{Painting$\rightarrow$}} && \multicolumn{3}{c}{\textbf{Sketches}$\rightarrow$} && \\
        \cline{2-4} \cline{6-8} \cline{10-12} \cline{14-16}
        Target & {C} & {P} & {S} && {R} & {P} & {S} && {R} & {C} & {S} && {R} & {C} & {P} && Avg. \\
        \midrule
        Resnet-50 & 54.60  & 57.92 & 43.71 && 50.87 & 38.37 & 43.92 && 66.65 & 50.33 & 39.87 && 48.28 & 52.46 & 43.47 && 49.20 \\
        MCD~\cite{saito2018maximum} & 52.94  & 57.29 & {40.38} && 55.71 & 43.69 & 47.57 && {67.80} & 51.88 & 44.95 && 56.83 & 56.32 & 50.83 && 52.18 \\
        RSDA~\cite{gu2020spherical} & 54.60 & 61.54 & 50.94 && 56.56 & 45.50 & 48.63 && 60.41 & 45.74 & 48.64 && 58.62 & 56.09 & 54.00 && 53.44 \\
        DANN~\cite{DANN} & 61.67 & 60.27 & 53.86 && 58.23 & 46.46 & 51.63 && 64.17 & 52.70 & 52.88 && 61.55 & 62.73 & 56.70 && 56.90  \\
        BSP~\cite{chen2019transferability} & 55.16  & 60.80 & 48.60 && 58.73 & 45.66 & \underline{55.47} && 65.18 & 48.59 & 48.58 && 61.40 & 56.78 & 55.79 && 55.06 \\
        SAFN~\cite{xu2019larger} & 55.81  & 64.82 & 48.50 && 58.68 & 49.96 & 52.42 && \textbf{73.71} & 56.25 & 53.54 && 64.32 & 60.65 & 59.53 && 58.18 \\
        CDAN~\cite{CDAN} & \underline{70.41}  & \underline{66.87} & \underline{57.73} && 61.61 & 50.90 & 54.72 && 68.47 & \underline{59.43} & \underline{55.49} && 64.27 & 64.22 & 59.14 && 61.11 \\
        PAN~\cite{PAN}$^\dagger$ & 67.56  & 66.73 & 55.86 && \underline{65.16} & \textbf{58.87} & 54.55 && 70.46 & 57.54 & 53.14 && \underline{66.55} & \underline{64.40} & \underline{60.22} && \underline{61.75} \\
        \midrule
        \Ours{} & \textbf{73.23}$^{\pm0.09}$ &	\textbf{70.46}$^{\pm{0.13}}$ &	\textbf{61.51}$^{\pm0.08}$ && \textbf{66.51}$^{\pm0.21}$ &	\underline{53.61}$^{\pm0.39}$ &	\textbf{58.79}$^{\pm0.68}$ && \underline{71.23}$^{\pm0.20}$ &	\textbf{63.17}$^{\pm0.75}$ &	\textbf{58.11}$^{\pm0.63}$	&& \textbf{67.60}$^{\pm0.16}$ &	\textbf{68.77}$^{\pm0.52}$ &	\textbf{64.09}$^{\pm0.51}$ && \textbf{64.76} \\ 
        \bottomrule
        \end{tabular}}
    \end{center}
    \caption{Accuracy scores on 126 classes on DomainNet. \textbf{Bold} and \underline{underline} indicate the best and next best methods respectively. $^\dagger{}$Uses hierarchical label annotation. }
    \label{tab:domainnet126}
\end{table*}


\vspace{-14pt}
\section{Using a different adaptation backbone}

The benefits obtained by \Ours{} are complementary to the nature of adaptation method used. In \tabref{tab:net+dann_sup} on \dnet{} dataset with 345 classes, we show gains starting from a DANN \cite{DANN} and CAN~\cite{kang2019contrastive} objective as well, {besides gains using the CDAN objective showcased in the main paper} using a Resnet-50 backbone. These results indicate that the benefits using our objective are available to a wide variety of alignment methods.

\begin{table*}[!!t]
    \begin{center} 
    
    \begin{subtable}[t]{0.95\hsize}\centering
    \setlength{\tabcolsep}{3pt}
    \resizebox{\textwidth}{!}{
    \begin{tabular}{@{} l *{17}{c} @{} }
        \toprule
        Source & \multicolumn{3}{c}{\textbf{Real$\rightarrow$}} && \multicolumn{3}{c}{\textbf{Clipart$\rightarrow$}} && \multicolumn{3}{c}{\textbf{Painting$\rightarrow$}} && \multicolumn{3}{c}{\textbf{Sketches}$\rightarrow$} && \\
        \cline{2-4} \cline{6-8} \cline{10-12} \cline{14-16}
        Target & {C} & {P} & {S} && {R} & {P} & {S} && {R} & {C} & {S} && {R} & {C} & {P} && Avg. \\
        \midrule
        DANN~\cite{DANN} & 45.93& 44.51& 35.47& & 46.85& 30.52& 36.77& & 48.02& 34.76& 32.15& & 47.1& 46.45& 38.47& & 40.58 \\
        DANN + \Ours{} & 49.67 & 48.61 & 39.14 & & 49.81 & 35.1 & 40.59 & & 50.04 & 38.51 & 36.61 & & 50.31 & 50.8 & 42.73 & & \textbf{44.32} \\
        \midrule
        CAN~\cite{kang2019contrastive} & 40.71 &37.77 &33.7 && {54.93} &31.41 &37.37 && 51.05 &33.64 &30.95 && {52.13} &42.19 &32.04 && 39.82 \\
        CAN + \Ours{} & 43.79 & 38.99 & 36.71 && 55.36 & 32.41 & 39.46 && 52.48 & 35.21 & 32.89 && 54.15 & 44.60 & 33.02 && \textbf{41.59} \\
        \bottomrule
        \end{tabular}}
        \captionsetup{width=\textwidth, font=footnotesize}
        \subcaption{Accuracy values of \Ours{} using DANN and CAN adaptation backbones on \dnet{}-345 classes. Note improved accuracy using \Ours{} on top of both the backbones.}
        \label{tab:net+dann_sup}
    \end{subtable}
    \vspace{4pt}

    \begin{subtable}[t]{0.95\hsize}\centering
    \setlength{\tabcolsep}{3pt}
    \resizebox{\textwidth}{!}{
    \begin{tabular}{@{} c *{17}{c} @{} }
        \toprule
        Source & \multicolumn{3}{c}{\textbf{Real$\rightarrow$}} && \multicolumn{3}{c}{\textbf{Clipart$\rightarrow$}} && \multicolumn{3}{c}{\textbf{Painting$\rightarrow$}} && \multicolumn{3}{c}{\textbf{Sketches}$\rightarrow$} && \\
        \cline{2-4} \cline{6-8} \cline{10-12} \cline{14-16}
        Target & {C} & {P} & {S} && {R} & {P} & {S} && {R} & {C} & {S} && {R} & {C} & {P} && Avg. \\
        \midrule
        Resnet-101 & 45.62 & 44.24 & 33.12 && 41.96 & 27.07 & 33.07 && 48.54 & 34.92 & 29.84 && 35.87 & 42.64 & 28.01 && 37.07 \\
        DANN~\cite{DANN} & 47.71 & 44.1 & 35.99 && 48.33 & 32.00 & 38.54 && 48.13 & 34.57 & 34.23 && 48.19 & 48.56 & 39.67 &&	41.67 \\
        MCD~\cite{saito2018maximum} & 41.11 &39.01 &26.1 && 40.77 &28.26 &33.02 && 45.49 &33.03 &29.1 && 38.29 &42.3 &29.51 && 35.49 \\
        CDAN~\cite{CDAN} & 52.47 & 48.0 & 40.42 & & 46.63 & 32.42 & 39.18 & & 48.81 & 37.92 & 35.39 & & 45.69 & 48.92 & 37.31 & & 42.76   \\
        SAFN~\cite{xu2019larger} & 44.93 &46.52 &28.2 && 37.2 &31.11 &36.3 && 53.32 &36.95 &32.48 && 44.12 &53.46 &40.05 && 40.38 \\
        ToAlign~\cite{wei2021toalign} & 50.10 & 48.27 & 35.98 && 50.24 & 31.41 & 41.10 && 54.60 & 43.67 & 36.82 && 50.15 & 54.32 & 42.06 && 44.89 \\
        \Ours{} & \textbf{56.25} & \textbf{52.96} & \textbf{42.22} & & \textbf{53.52} & \textbf{37.46} & \textbf{43.46} & & \textbf{53.38} & \textbf{42.69} & \textbf{39.65} & & \textbf{53.17} & \textbf{55.29} & \textbf{44.29} & & \textbf{47.86} \\
        \bottomrule
        \end{tabular}}
        \subcaption{Results on \dnet{}-345 dataset with Resnet-101 backbone and {batch size of 24}. }
        \label{tab:resnet101_supp}
    \end{subtable}
    
    \end{center}
    \vspace{-16pt}
    \caption{{\bf Ablations on \dnet{}-345 dataset}.}
\end{table*}

\section{\Ours{} with deeper ResNets}

In \tabref{tab:resnet101_supp}, we show the results of the baselines as well as \Ours{} with a deeper Resnet-101 backbone. \textbf{Due to memory constraints, we use a batch size of 24 for all the methods (unlike the experiments with Resnet-50 in the main paper where we use a batch size of 32).} It is clearly evident that relative improvements by \Ours{} over other baselines are still significant, indicating that our benefits persist even with a more powerful CNN backbone.

\section{Ablations on KNN-based pseudo-labeling}
\label{sec:knnpseudo}

\begin{figure*}[!!h]
    \begin{minipage}[b]{0.45\textwidth}
        \centering
        \includegraphics[width=\textwidth]{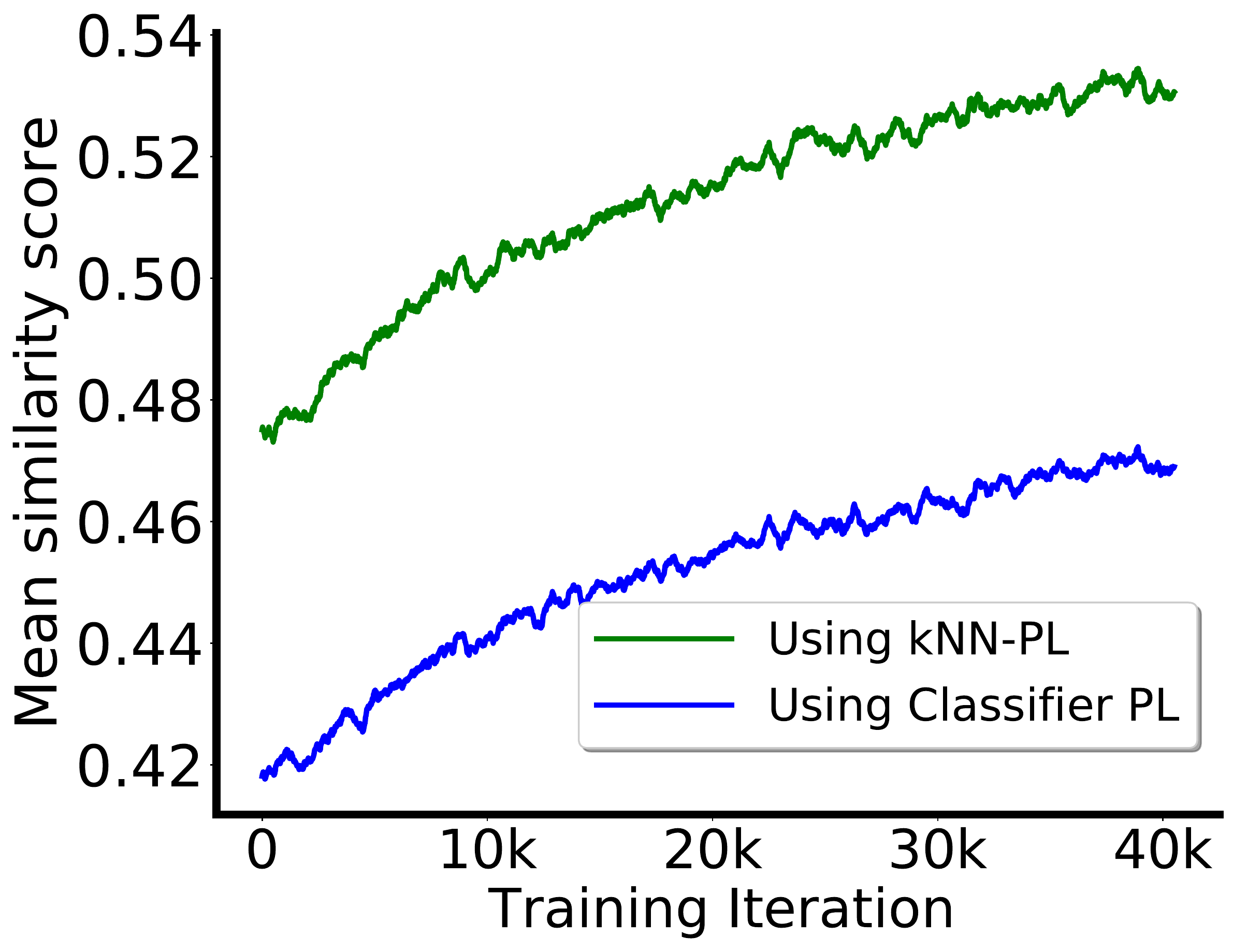}
        \vspace{-18pt}
        \caption{Similarity Score}
        \label{fig:sim_pl}
    \end{minipage}
    \hfill
    \begin{minipage}[b]{0.45\textwidth}
        \centering
        \includegraphics[width=\textwidth]{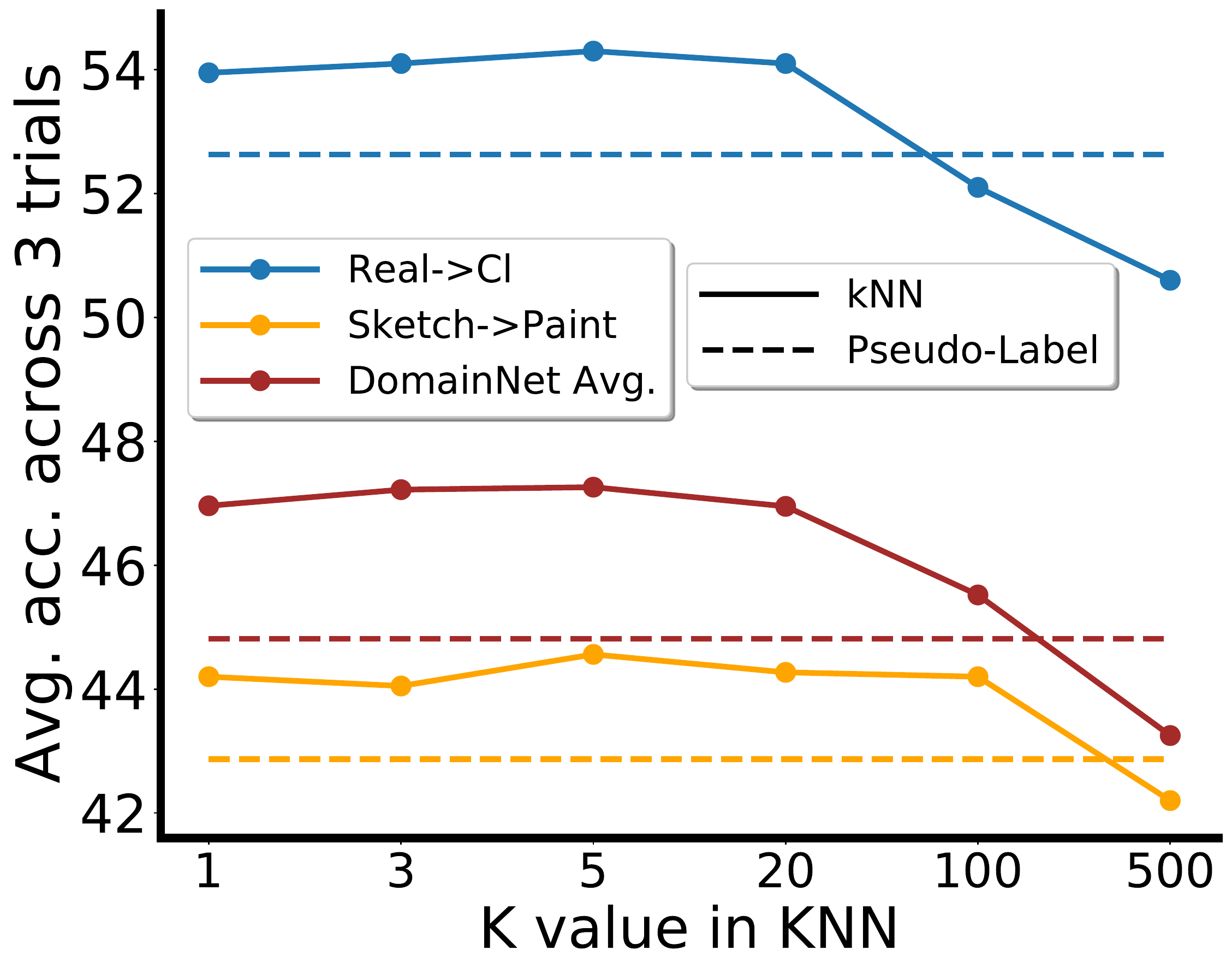}
        \vspace{-18pt}
        \caption{Effect of K}
        \label{fig:classpl}
    \end{minipage}
    \vspace{-12pt}
\end{figure*}

A crucial choice made in the design of \Ours{} is the use of kNN-based pseudo-labeling instead of directly using the classifier predictions on unlabeled target samples as pseudo-labels for all the target samples. This follows from the observation that the kNN based pseudo-labeling is generally robust to noisy classifier boundaries, especially amidst domain shifts. Moreover, with the help of the memory bank, the neighborhood from which the nearest neighbors are computed is much larger than the size of the mini-batch. We verify this intuition in \figref{fig:sim_pl}, where the mean similarity score between the samples from the same class is much higher when trained using the proposed kNN based pseudo-labeling technique as compared to the classifier based pseudo-labeling technique.  Furthermore, we analyze the effect of the choice of the parameter K in \figref{fig:classpl}. Our accuracy is robust to most values of K in the range of 1-20. At large values of K, however, the accuracy falls steeply due to large amounts of noise in the pseudo-labels. Additionally, in \tabref{tab:memknn}, we show that both the memory bank and kNN based pseudo-labeling are crucial to achieve performance gains using the consistency loss, as removing one of them (or both of them) results in significant drop in performance.

\section{Results on Office-Home dataset}

In the main paper, we show results using largest available datasets for domain adaptation, namely \dnet{}-345 as well as CUB-200 with 345 and 200 categories, respectively. In \tabref{tab:officehome}, we show result using a medium-sized dataset, Office-Home~\cite{venkateswara2017deep}. Office-home contains 65 categories across 4 domains, and around 4k images in each domain. We observe that \Ours{} outperforms competitive baselines even on Office-Home. Specifically, we use CDAN as the adaptation backbone, and report an improvement of $1.12\%$ over this baseline, indicating the effectiveness of \Ours{} even for unsupervised adaptation even on medium-sized datasets.  

\begin{figure*}[!h]
     \centering
     \begin{subfigure}[t]{0.45\textwidth}
        \centering
        \includegraphics[width=\textwidth]{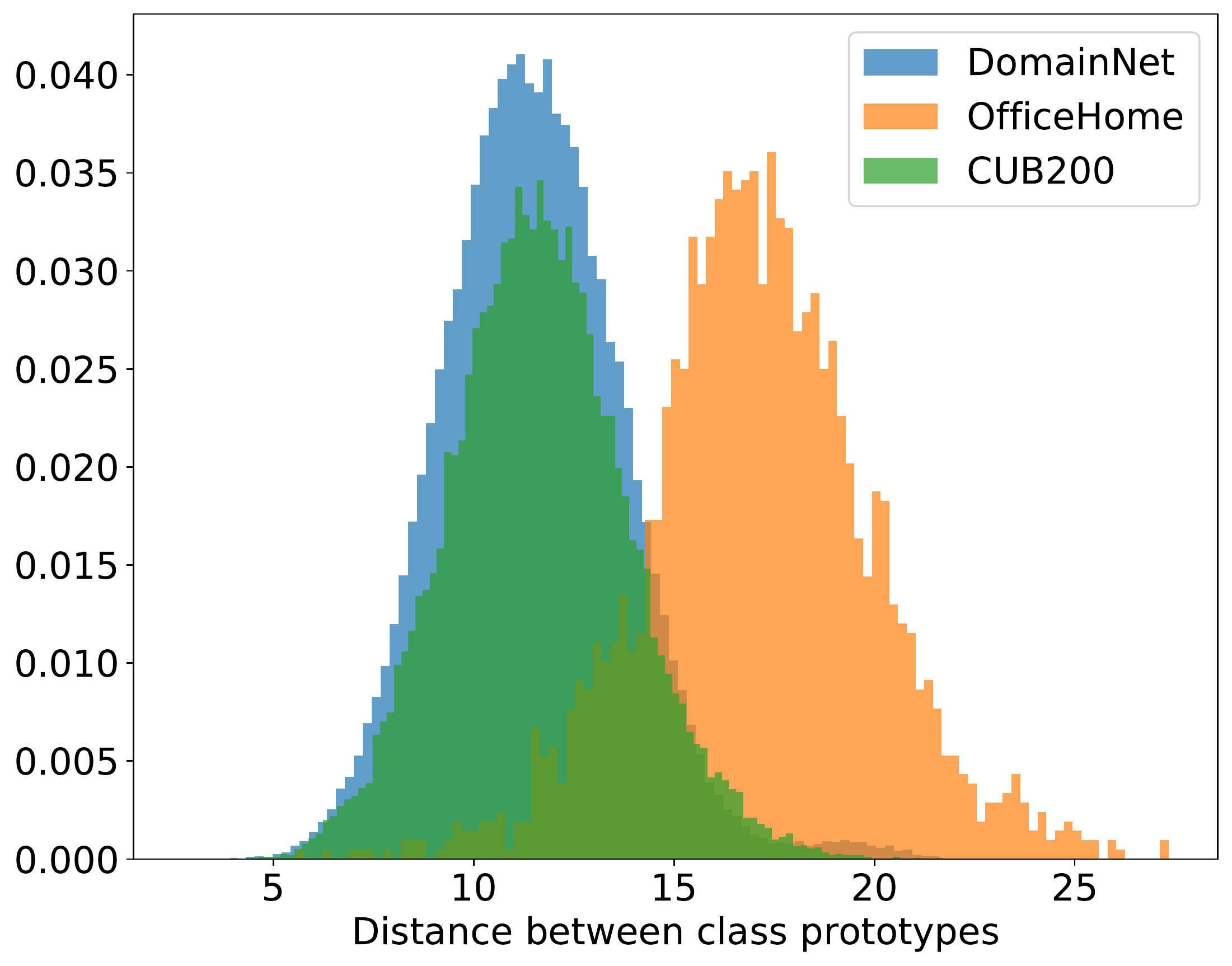}
        \captionsetup{width=0.95\textwidth, font=footnotesize}
        \subcaption{Histogram of distances between class prototypes from each dataset. Lesser distance indicates more class confusion and more difficulty.}
        \label{fig:distances}
     \end{subfigure}
     \hfill
     \begin{subfigure}[t]{0.45\textwidth}
        \centering
        \includegraphics[width=\textwidth]{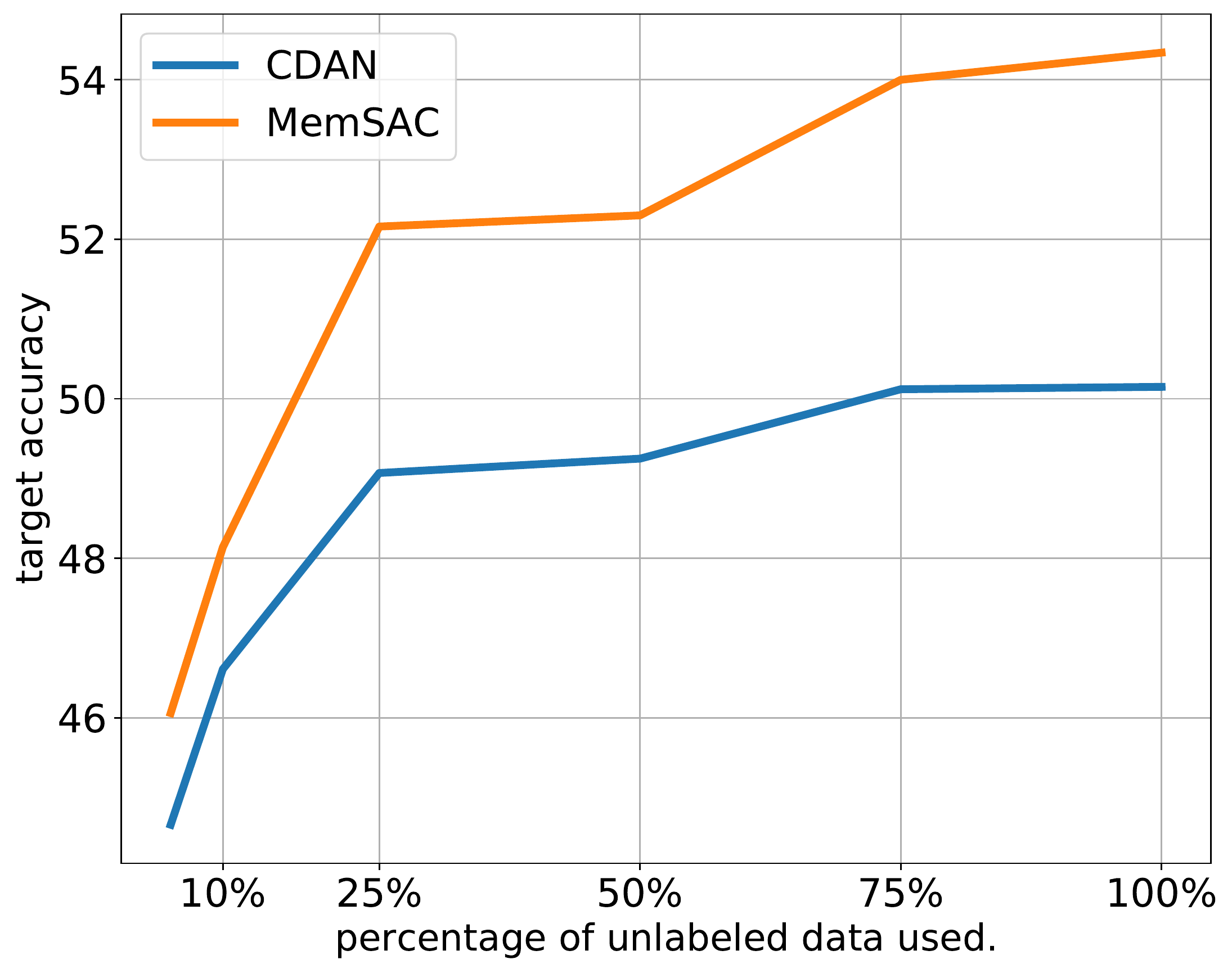}
        \captionsetup{width=0.98\textwidth, font=footnotesize}
        \subcaption{Accuracy using varying degrees of unlabeled samples from \dnet{}-345 dataset on \textbf{R} $\rightarrow$ \textbf{C}.}
        \label{fig:unlabeled}
     \end{subfigure}
     \caption{(\subref{fig:distances}) shows the distribution of distances between class prototypes from each category from \dnet{}, CUB-200 and OfficeHome datasets. \dnet{} and CUB-200 have lesser inter-class distances than Office-Home. (\subref{fig:unlabeled}) shows the effect of using varying degree of target unlabeled samples on the adaptation performance. The performance of \Ours{} consistently improves as more unlabeled data becomes available. }
\end{figure*}

However, the results of \Ours{} on OfficeHome dataset are not SOTA, which might be attributed to two reasons. Firstly, the categories in OfficeHome dataset are clearly distinct from each other leading to little avenues for negative alignment, which \Ours{} tries to alleviate. To illustrate this, we use a ImageNet pre-trained Resnet-50 model and compute feature embeddings for all images from \dnet{}-clipart domain. We then use the ground truth labels to find the class prototypes (or per-class average feature) for all the 345 classes and compute pairwise Euclidean distance between class prototypes. Lower euclidean distances between class prototypes indicates more class confusion and more likely negative transfer. We compute inter-class distances for OfficeHome and CUB200 datasets as well in similar fashion and show results in \figref{fig:distances}. Evidently, the inter-class distances between classes from DomainNet and CUB200 are much smaller compared to OfficeHome, and hence greater benefit in using an approach like \Ours{}.

\begin{table*}[!t]
    \centering
    \captionsetup{width=\textwidth, font=footnotesize}
    \caption{Accuracy scores on 65 categories on OfficeHome~\cite{venkateswara2017deep} dataset. }
    \label{tab:officehome}
    \setlength{\tabcolsep}{3pt}
    
    \resizebox{\textwidth}{!}{
\begin{tabular}{lccccccccccccc}
    \toprule
    Method & \small{$\mathbf{A} {\rightarrow} \mathbf{C}$} & \small{$\mathbf{A} {\rightarrow} \mathbf{P}$} & \small{$\mathbf{A} {\rightarrow} \mathbf{R}$} & \small{$\mathbf{C} {\rightarrow} \mathbf{A}$} & \small{$\mathbf{C} {\rightarrow} \mathbf{P}$} & \small{$\mathbf{C} {\rightarrow} \mathbf{R}$} & \small{$\mathbf{P} {\rightarrow} \mathbf{A}$} & \small{$\mathbf{P} {\rightarrow} \mathbf{C}$} & \small{$\mathbf{P} {\rightarrow} \mathbf{R}$} & \small{$\mathbf{R} {\rightarrow} \mathbf{A}$} &  \small{$\mathbf{R} {\rightarrow} \mathbf{C}$} & \small{$\mathbf{R} {\rightarrow} \mathbf{P}$} & AVG \\
    \midrule
    Resnet-50 & 34.9 & 50.0 & 58.0 & 37.4 & 41.9 & 46.2 & 38.5 & 31.2 & 60.4 & 53.9 & 41.2 & 59.9 & 46.1 \\
    \midrule
    DANN~\cite{DANN} & 45.6 & 59.3 & 70.1 & 47.0 & 58.5 & 60.9 & 46.1 & 43.7 & 68.5 & 63.2 & 51.8 & 76.8 & 57.6 \\
    JAN~\cite{long2017deep} & 45.9 & 61.2 & 68.9 & 50.4 & 59.7 & 61.0 & 45.8 & 43.4 & 70.3 & 63.9 & 52.4 & 76.8 & 58.3 \\
    CDAN~\cite{CDAN} & 50.7 & 70.6 & 76.0 & 57.6 & 70.0 & 70.0 & 57.4 & 50.9 & 77.3 & 70.9 & 56.7 & 81.6 & 65.8 \\
    BSP~\cite{chen2019transferability} & 52.0 & 68.6 & 76.1 & 58.0 & {70.3} & 70.2 & 58.6 & 50.2 & {77.6} & {72.2} & {59.3} & {81.9} & 66.3\\
    SAFN~\cite{xu2019larger} & {52.0} & {71.7} & {76.3} & {64.2} & 69.9 & {71.9} & {63.7} & {51.4} & 77.1 & 70.9 & 57.1 & 81.5 & {67.3} \\
    RSDA~\cite{gu2020spherical} & 53.2 & 77.7 & 81.3 & 66.4 & 74.0 & 76.5 & 67.9 & 53.0 & 82.0 & 75.8 & 57.8 & 85.4 & \textbf{70.9} \\
    \Ours{} & {53.10} & {73.7} & {77.8} & {62.9} & {71.22} & {72.32} & {61.22} & {51.93} & {79.22} & {75.0} & {59.39} & {83.35} & {68.42} \\
    \bottomrule
    \end{tabular}}

\end{table*}

A second reason could be that the amount of unlabeled data in OfficeHome is much lesser compared to DomainNet on any transfer setting. On average, around 30k unlabeled samples are available in any target domains from DomainNet while only 4k samples are available from OfficeHome. To verify this argument, we subsample the \textit{clipart} domain from \dnet{} to only use \{5,10,25,50,75\}\% of unlabeled data during adaptation on the transfer task \textbf{R} $\rightarrow$ \textbf{C}. As indicated in \figref{fig:unlabeled}, the benefits from \Ours{} grows significantly when larger unlabeled data is available. Note that OfficeHome contains only 10\% of unlabeled data compared to \dnet{}, and from \figref{fig:unlabeled}, the gains using \Ours{} is minimal.




\section{Category wise accuracies on \dnet{}}

\begin{figure*}[tbp]
     \centering
     \begin{subfigure}[t]{0.45\textwidth}
        \centering
        \includegraphics[width=\textwidth]{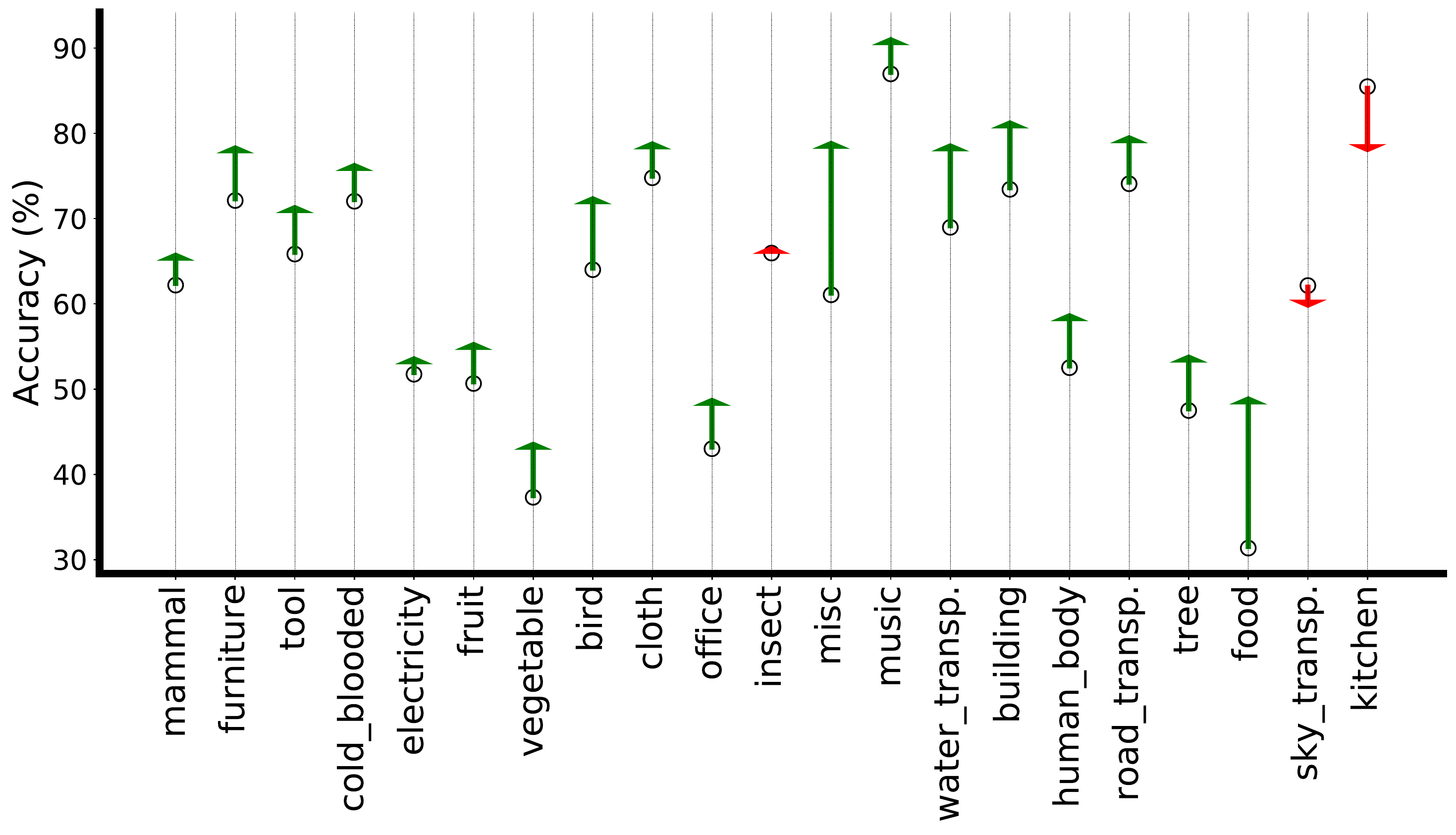}
        \captionsetup{width=0.95\textwidth, font=footnotesize}
        \subcaption{\textbf{C} $\rightarrow$ \textbf{R}}
        \label{fig:dnet_cr}
     \end{subfigure}
     \hfill
    \begin{subfigure}[t]{0.45\textwidth}
        \centering
        \includegraphics[width=\textwidth]{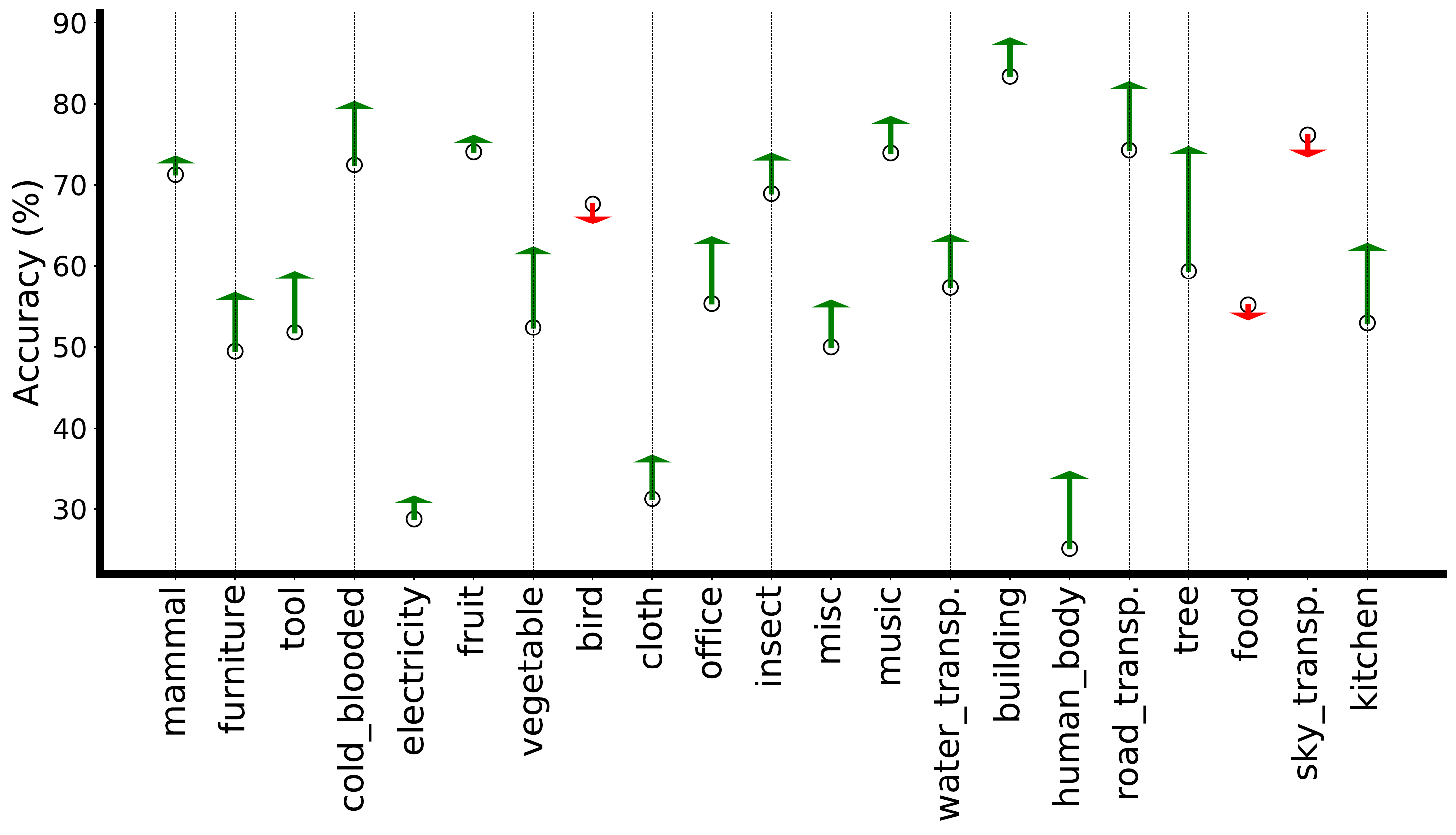}
        \captionsetup{width=0.98\textwidth, font=footnotesize}
        \subcaption{\textbf{P} $\rightarrow$ \textbf{C}}
        \label{fig:dnet_pc}
     \end{subfigure}
     
     \begin{subfigure}[t]{0.45\textwidth}
        \centering
        \includegraphics[width=\textwidth]{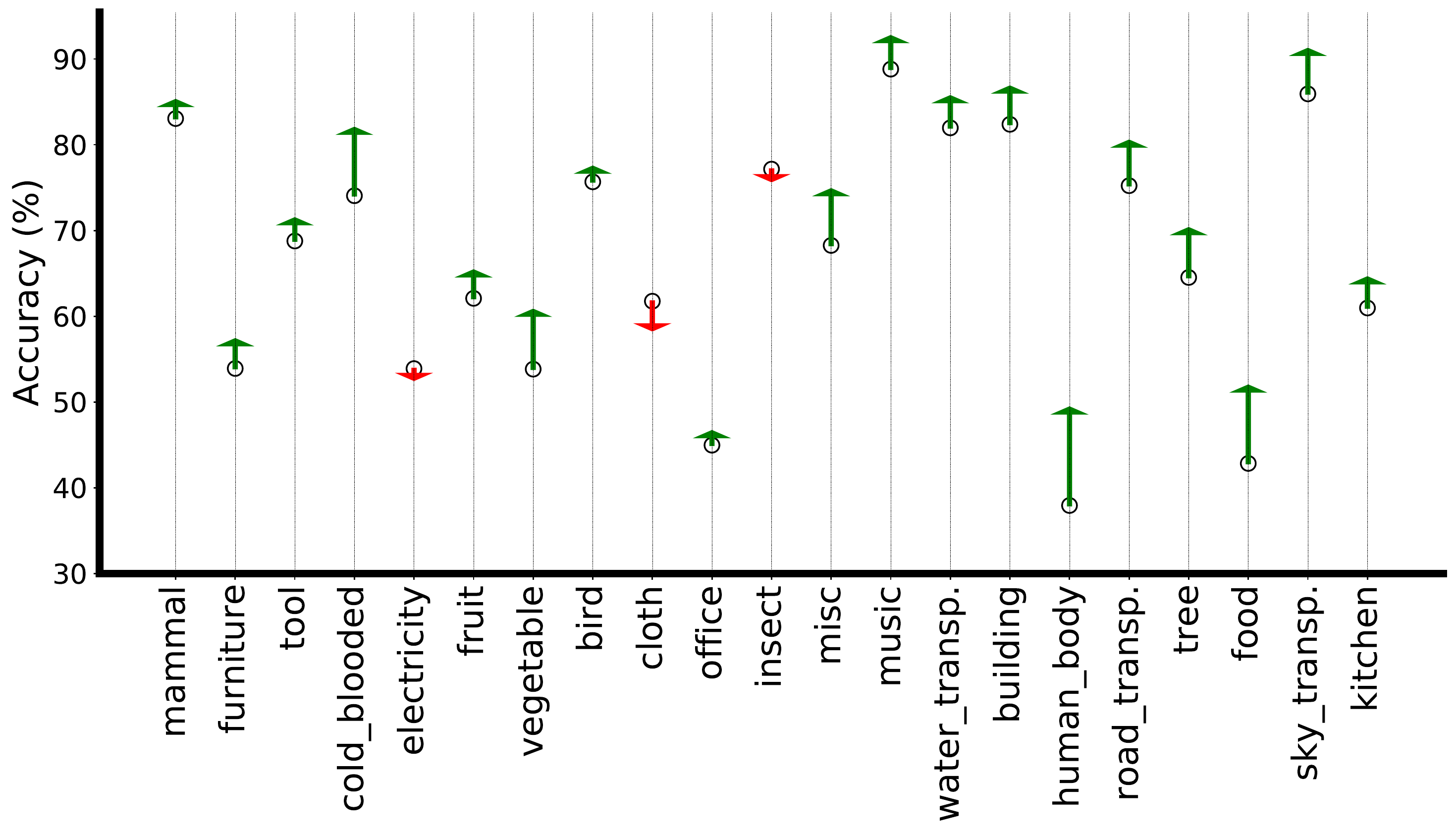}
        \captionsetup{width=0.95\textwidth, font=footnotesize}
        \subcaption{\textbf{P} $\rightarrow$ \textbf{R}}
        \label{fig:dnet_pr}
     \end{subfigure}
     \hfill
     \begin{subfigure}[t]{0.45\textwidth}
        \centering
        \includegraphics[width=\textwidth]{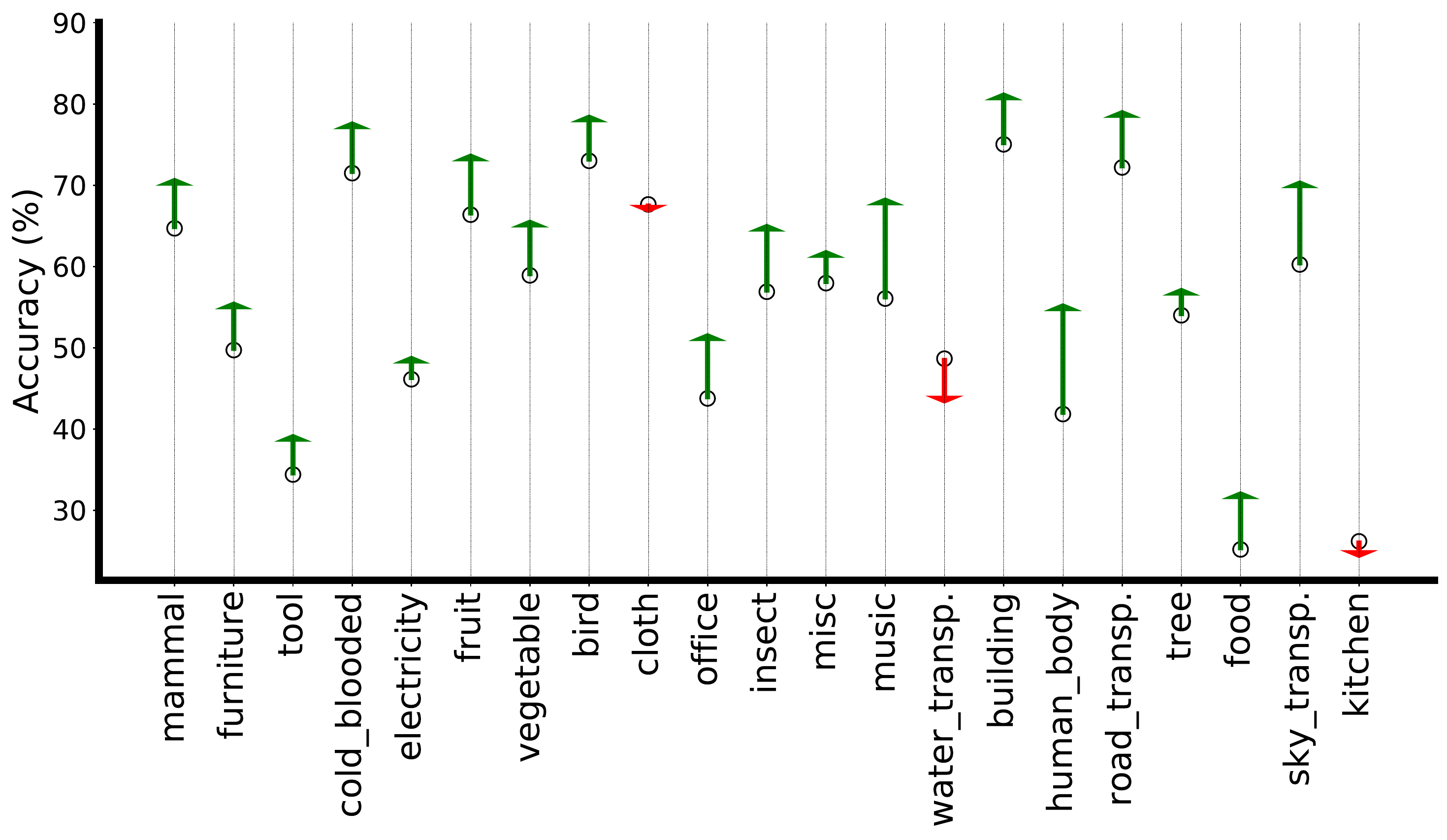}
        \captionsetup{width=0.95\textwidth, font=footnotesize}
        \subcaption{\textbf{S} $\rightarrow$ \textbf{P}}
        \label{fig:dnet_sp}
     \end{subfigure}
     \caption{Category wise accuracy increase and decrease on \dnet{} dataset compared with CDAN baseline.}
     \label{fig:cateWise_dnet_supp}
\vspace{-0.3cm}
\end{figure*}
We show the accuracy for each \textit{coarse} category and the gain/fall in accuracy between the baseline CDAN and \Ours{} in \figref{fig:cateWise_dnet_supp} for few more tasks from \dnet{}, in addition to the R$\rightarrow$C task shown in the main paper. Evidently, \Ours{} has non-trivial benefits over the baseline over most of the categories (marked by \textcolor{green}{$\uparrow$}), and any drops in accuracy (marked by \textcolor{red}{$\downarrow$}) are negligible. For example, on the task of P$\rightarrow$C, we observe improvements of 14.6\% on \texttt{trees} and 4.2\% on category \texttt{insects}, thus indicating that our benefits sustain over most categories, and are more pronounced for categories containing finer grained classes. 


\section{Results on Birds-123 and CompCars datasets}
\label{sec:compcars}

In addition to the results on CUB200 in the main paper for fine-grained adaptation, we also show the results using \Ours{} on other fine-grained datasets such as Birds-123 and CompCars~\cite{yang2015large}. Birds-123 contains images of different bird species from 123 common categories across CUB, NaBirds and iNaturalist datasets. CompCars, on the other hand, contains images from web and surveillance domains of 181 car models, and involves domain shift in the form of curated web images vs in-the-wild surveillance footage. Our method efficiently handles the domain shift across these challenging settings as shown in \tabref{tab:fgdata}. \Ours{} attains an accuracy of 78.42\% on the Birds-123 dataset and 52.75\% on CompCars dataset which is much higher than all prior methods including PAN, even though PAN is specifically designed for fine-grained adaptation. These results verify the effectiveness of \Ours{} on challenging fine-grained dataset settings.

\begin{table*}[ht]
    \begin{minipage}[b]{0.3\textwidth}
    \centering
        \resizebox{\textwidth}{!}{
        \begin{tabular}{@{} l *{3}{c} @{} }
            \toprule
            & {Birds-123} & {CompCars}\\
            \midrule
            Source & 72.02 & 15.64 \\
            DANN & 63.36 & 48.90 \\
            PAN & 74.04 & 48.62 \\
            CDAN & 72.95 & 50.40 \\
            MemSAC & 78.42 & 52.75 \\
            \bottomrule
        \end{tabular}
        }
        \captionsetup{width=\textwidth, font=footnotesize}
        \subcaption{}
        \label{tab:fgdata}
    \end{minipage}
    \hfill
    \begin{minipage}[b]{0.3\textwidth}
        \centering
        \resizebox{\textwidth}{!}{
        \begin{tabular}{@{} l *{3}{c} @{} }
            \toprule
            & \footnotesize{W/ kNN} & \footnotesize{Classifier PL}\\
            \midrule
            w/ Mem. & \textbf{47.26} & 44.81 \\
            w/o Mem. & 43.32 & 43.24 \\
            \bottomrule
        \end{tabular}
        }
        \captionsetup{width=\textwidth, font=footnotesize}
        \subcaption{} 
        \label{tab:memknn}
    \end{minipage}
    \hfill
    \begin{minipage}[b]{0.3\textwidth}
        \centering
        \resizebox{\textwidth}{!}{
        \begin{tabular}{@{} l *{3}{c} @{} }
            \toprule
            & DomainNet \\
            \midrule
            Source Only & 35.98 \\
            + MemSAC & 38.11(\textcolor{blue}{+2.13\%}) \\
            \midrule
            CDAN & 43.24 \\
            + MemSAC & 47.26(\textcolor{blue}{+4.02\%}) \\
            \bottomrule
        \end{tabular}
        }
        \captionsetup{width=\textwidth, font=footnotesize}
        \subcaption{} 
        \label{tab:sourcemsac}
    \end{minipage}
    \caption{In \subref{tab:fgdata}, we show the comparison of \Ours{} with prior methods on fine-grained datasets Birds-123 and CompaCars. In \subref{tab:memknn}, we show the role of memory module and kNN pseudo labeling. In \subref{tab:sourcemsac}, we show the role of adversarial losses to improve \Ours{} training.}
    \vspace{-18pt}
\end{table*}

\section{Role of adaptation in \Ours{}}

We next verify the role of adaptation losses in our proposed framework. While \Ours{} can efficiently improve alignment using similarity consistency losses, we still need to bootstrap the training using adaptation losses for few iterations, to avoid noisy pseudo-labels in the later stages of training. As shown in \tabref{tab:sourcemsac}, while \Ours{} can still boost performance of \textit{Source Only} model, the gains observed using  \Ours{} alongside adaptation losses like CDAN are much higher.

\section{Queue updates using momentum encoder}
\label{sec:momentum_update}

We now discuss possible alternative strategies to update the memory bank. For this purpose, we generalize the update rule using a \textit{momentum encoder}, proposed in \cite{he2020momentum}. After the initial bootstrapping phase where we train the encoder on source data for few iterations, we initialize the momentum encoder $\F$ using the state of the encoder $\E$. After that, at every iteration, the parameters of the momentum encoder $\theta_{\mathcal{F}}$ are updated as follows.
\begin{equation}
    \theta_{\F} = (1-\mu) * \theta_{\E} + (\mu) * \theta_{\F}
    \label{eq:momentum_update}
\end{equation}
Here, $\mu$ is called the momentum parameter, and controls the speed of updates. The source features encoded in the memory bank $\M$ are obtained by a forward pass on $\F$, while the source features used to compute the supervised loss as well as all the target features are computed using a forward pass on $\E$. We note that the original update rule discussed in the main paper is just a special case of Eq.~\eqref{eq:momentum_update}, which is obtained by putting $\mu=0$. 

The intuition behind using such a momentum based encoder is that it gives features with a slow drift through the training, and hence can support larger queues. We use such a momentum update on \Ours{} and show results for CUB-Drawings dataset in \tabref{tab:momentumEffect} 
We found no benefit using such a momentum encoder in our method. This might be because we already bootstrap the encoder until the features stabilize and achieve a slow-drift phenomenon, and using momentum based updates on top of that might not improve accuracy. In light of these results, designing better memory bank update schedules is left as a potential direction for future work.

\section{Effect of loss coefficient}

\begin{table}[!tbp]
\centering
    \begin{subtable}[t]{0.4\hsize}\centering
        \resizebox{\textwidth}{!}{
        \begin{tabular}{@{} c c c c @{} }
            \toprule
            $\mu$ & C$\rightarrow$D & D$\rightarrow$C & Avg. Acc. \\
            \midrule
            0 & \textbf{73.97} & \textbf{61.94} & \textbf{67.95} \\
            0.5 & 68.61 & 55.24 & 61.92 \\
            0.9 & 68.89 & 55.24 & 62.06 \\
            0.999 & 71.43 & 58.81 & 65.12 \\
            \bottomrule
        \end{tabular}}
        \captionsetup{width=0.95\textwidth, font=footnotesize}
        \caption{\Ours{} with different values of momentum parameter $\mu$. }
        \label{tab:momentumEffect}
    \end{subtable}
    \hfill
    \begin{subtable}[t]{0.4\hsize}\centering
        \resizebox{\textwidth}{!}{
        \begin{tabular}{@{} c c c c @{} }
            \toprule
            $\lambda_{sc}$ & C$\rightarrow$D & D$\rightarrow$C & Avg. Acc. \\
            \midrule
            0.001 & 65.84 & 51.29 & 58.56 \\
            0.01 & 69.38 & 55.91 & 62.64 \\
            0.1 & \textbf{73.97} & \textbf{61.94} & \textbf{67.95} \\
            1 & 19.02 & 50.56 & 36.29 \\
            \bottomrule
        \end{tabular}}
        \captionsetup{width=0.95\textwidth, font=footnotesize}
        \caption{Effect of loss coefficient $\lambda_{sc}$ on the accuracy for CUB-Drawings dataset on 200 classes. }
        \label{tab:lossCoeff}
    \end{subtable}
    \caption{{\bf Ablation on CUB-Drawing dataset} using Resnet-50 backbone}
\end{table}
We show the ablation using the loss coefficient of our sample consistency loss in \tabref{tab:lossCoeff} on CUB-Drawing dataset. We find that using a value of $\lambda_{sc}$ as 0.1 gave the best result, while using any larger value gives much inferior results, as noisy negative and positive pairs have a high influence on the training.



\section{Training Curves}

\begin{figure*}
     \centering
     \begin{subfigure}[t]{0.32\textwidth}
        \centering
        \includegraphics[width=\textwidth]{figures/SimScores.pdf}
        \captionsetup{width=0.95\textwidth, font=footnotesize}
        \subcaption{}
        \label{fig:simscores_supp}
     \end{subfigure}
     \hfill
    \begin{subfigure}[t]{0.32\textwidth}
        \centering
        \includegraphics[width=\textwidth]{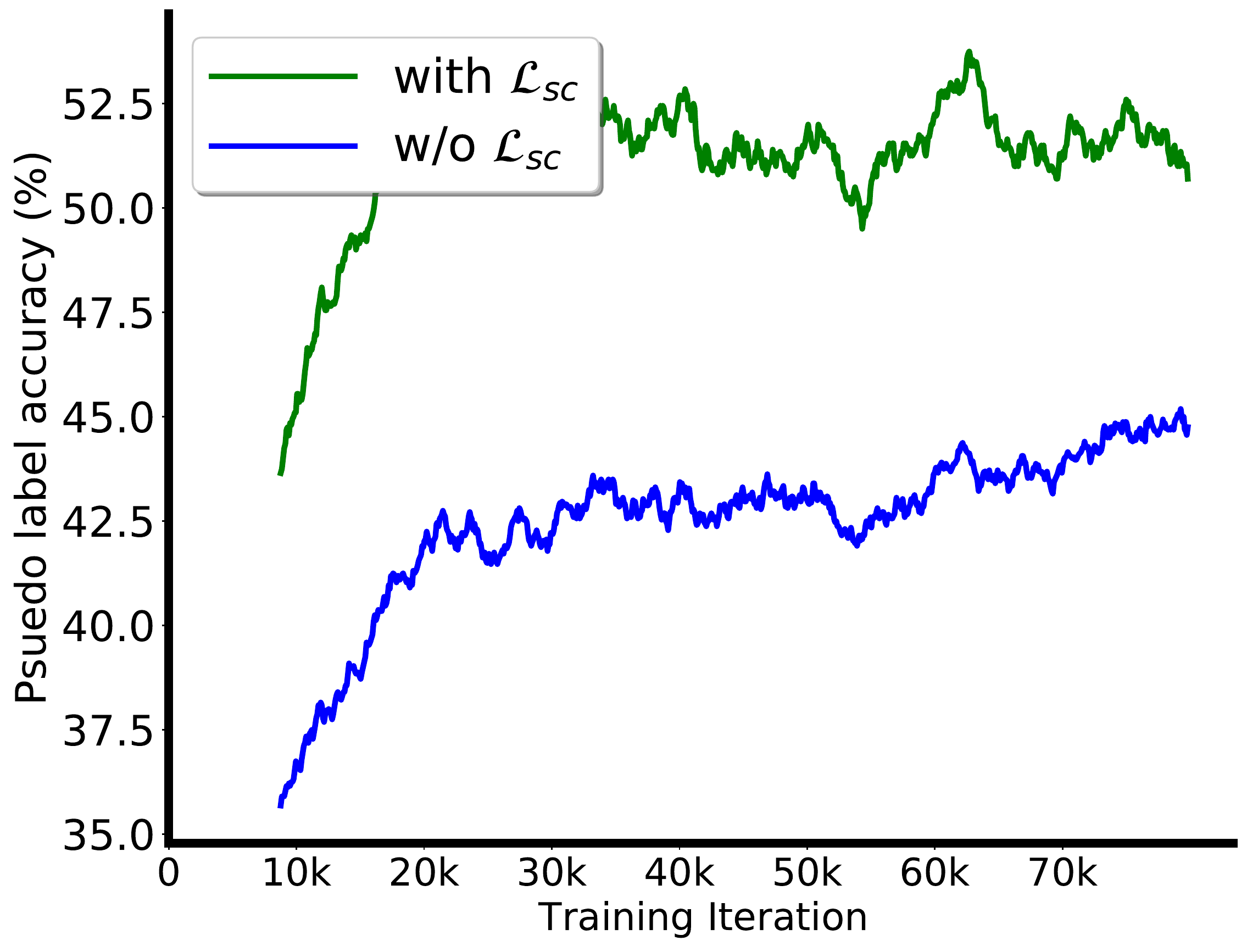}
        \captionsetup{width=0.98\textwidth, font=footnotesize}
        \subcaption{}
        \label{fig:pLacc_supp}
     \end{subfigure}
     \hfill
     \begin{subfigure}[t]{0.32\textwidth}
        \centering
        \includegraphics[width=\textwidth]{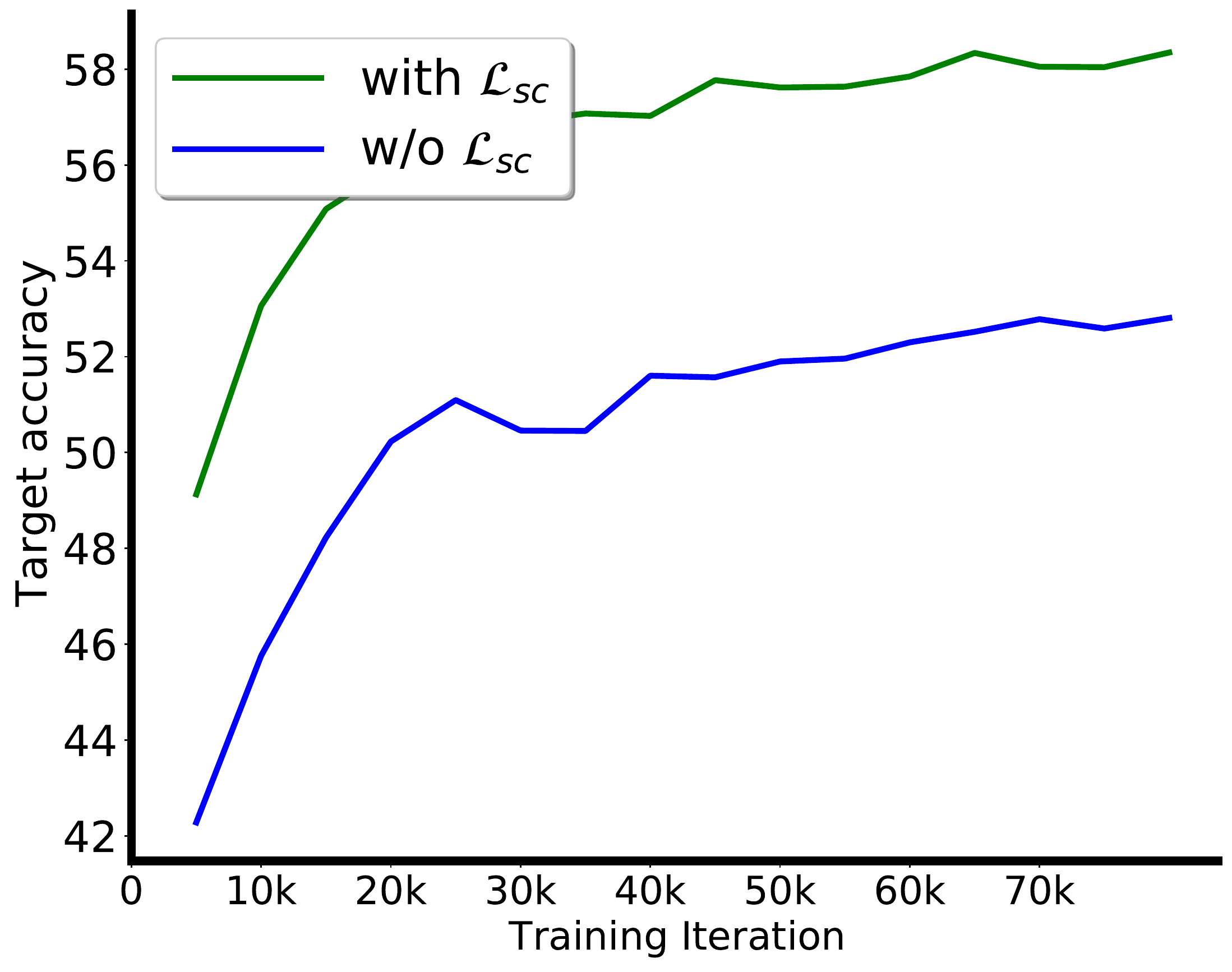}
        \captionsetup{width=0.95\textwidth, font=footnotesize}
        \subcaption{}
        \label{fig:tarAcc_supp}
     \end{subfigure}
     \caption{{\bf Training Curves for \textit{D} $\rightarrow$ \textit{C}}
     (\subref{fig:simscores_supp}) Mean similarity score of within class samples vs. Training iterations.
     (\subref{fig:pLacc_supp}) Pseudo-label accuracy vs. Training iterations.
     (\subref{fig:tarAcc_supp}) Final target accuracy vs. Training iterations
      }
     \label{fig:training_curves}
\end{figure*}

In \figref{fig:training_curves}, we show the trends for the mean similarity score, psuedo label accuracy as well as the final target accuracy during training. We compare between \Ours{} which uses a consistency based loss, with an approach which does not contain such a consistency constraint. We observe that using our sample consistency loss gives a higher value of mean similarity score, psuedo-label accuracy as well as final target accuracy during training, and each of them improve with training indicating the effectiveness of our proposed loss.

\section{Training details}

In \tabref{tab:hyperparams}, we give complete details regarding all the hyperparameters used for the experiments. While all the hyperparameters are same across both \dnet{} and CUB-Drawings, we use a memory bank $\M$ of size 24k for CUB-Drawings and 48k for DomainNet. This is because datasets with larger number of images can give benefit with larger memory banks. 

All the models were implemented using PyTorch 1.4.0 using 2080Ti GPUs. Following \cite{CDAN}, learning rate is $0.003$ for the feature encoder which is pretrained on ImageNet and $0.03$ for the classifier. 

\begin{table}[h]
  \centering
    \resizebox{0.3\textwidth}{!}{
    \begin{tabular}{l|c}
       Hyperparameter  & Value \\
       \midrule
    BatchSize & 32 \\
    QueueSize & 48000 \\
    $(\lambda_{adv} , \lambda_{sc})$ & (1,0.1)  \\
    Temperature $\tau$ & 0.07  \\
    Bootstrap Iter. & 4000  \\
    Total Iterations & 90k  \\
    k in kNN & 5 \\
    Learning Rate for $\E$ & 0.003 \\
    Learning rate for $\G$ and $\C$ & 0.03 \\
    No. of GPUs & 1\\
    \end{tabular}
        }
    \caption{Values of hyperparameters used in training \Ours{} on all the experiments.}
    \label{tab:hyperparams}
\end{table}

\section{Limitations}

Domain adaptation aims to efficiently address the problem of labeling overhead in low-resource domains enabling equitable performance of machine learning models across geographic, social or economic factors. However, \Ours{} shares with other deep domain adaptation approaches the limitation of lack of explainability and uncalibrated model uncertainty, which may have a negative impact on applications where decisions based on domain adaptation have a bearing on safety or equity. Moreover, we also note the significant room for improvement to achieve accuracy levels of fully supervised models, as noted in Table 1 in the main paper (\Ours{} vs. Tgt. Supervised). 

\end{appendix}
\end{document}